\documentclass[10pt,twocolumn,letterpaper]{article}
\usepackage{cvpr}      

\usepackage{graphicx}
\usepackage{amsmath}
\usepackage{amssymb}
\usepackage{booktabs}
\usepackage{makecell}
\usepackage[toc,page]{appendix}
\usepackage{multirow}
\newcolumntype{x}[1]{>{\centering\let\newline\\\arraybackslash\hspace{0pt}}p{#1}}
\usepackage{xcolor}
\makeatletter
\DeclareRobustCommand{\itshape}{%
  \not@math@alphabet\itshape\mathit
  \fontshape\itdefault\selectfont
  \color{gray}%
}
\makeatother
\usepackage{algorithm}
\usepackage{algpseudocode}
\usepackage{pifont}
\usepackage{listings}
\usepackage{etoolbox}
\usepackage{lipsum}
\usepackage{mathtools}
\usepackage{verbatim}

\usepackage[pagebackref,breaklinks,colorlinks]{hyperref}
\definecolor{commentgreen}{rgb}{0.0, 0.27, 0.13}

\usepackage[capitalize]{cleveref}
\crefname{section}{Sec.}{Secs.}
\Crefname{section}{Section}{Sections}
\Crefname{table}{Table}{Tables}
\crefname{table}{Tab.}{Tabs.}

\def\cvprPaperID{..} 
\def\confName{Under Review ..}

\begin{document}

\title{MC-SSL0.0: Towards Multi-Concept Self-Supervised Learning}

\author{Sara Atito\\
{\tt\small sara.atito@gmail.com}
\and
Muhammad Awais\\
{\tt\small m.a.rana@surrey.ac.uk}
\and
Ammarah Farooq\\
{\tt\small  ammarah.farooq@surrey.ac.uk}
\and
Zhenhua Feng\\
{\tt \small  z.feng@surrey.ac.uk}
\and
Josef Kittler\\
{\tt \small  j.kittler@surrey.ac.uk} \\
\and \hspace*{0cm}
Centre for Vision, Speech and Signal Processing, University of Surrey, Guildford\\
}
\maketitle

\begin{abstract}
Self-supervised pretraining is the method of choice for natural language processing models and is rapidly gaining popularity in many vision tasks. Recently, self-supervised pretraining has shown to outperform supervised pretraining for many downstream vision applications, marking a milestone in the area. This superiority is attributed to the negative impact of incomplete labelling of the training images, which convey multiple concepts, but are annotated using a single dominant class label. Although Self-Supervised Learning (SSL), in principle, is free of this limitation, the choice of pretext task facilitating SSL is perpetuating this shortcoming by driving the learning process towards a single concept output. This study aims to investigate the possibility of modelling all the concepts present in an image without using labels. In this aspect the proposed SSL framework MC-SSL0.0 is a step towards Multi-Concept Self-Supervised Learning (MC-SSL) that goes beyond modelling single dominant label in an image to effectively utilise the information from all the concepts present in it. MC-SSL0.0 consists of two core design concepts, group masked model learning and learning of pseudo-concept for data token using a momentum encoder (teacher-student) framework. The experimental results on multi-label and multi-class image classification downstream tasks demonstrate that MC-SSL0.0 not only surpasses existing SSL methods but also outperforms supervised transfer learning. The source code will be made publicly available for community to train on bigger corpus. \footnote{Under Review .....}

\end{abstract}

\begin{figure}[t!]
    \centering
    
    \begin{subfigure}[t]{\linewidth}
    \centering
        \begin{subfigure}[t]{0.23\linewidth}
        \caption*{Original Images}
        \includegraphics[width=\textwidth]{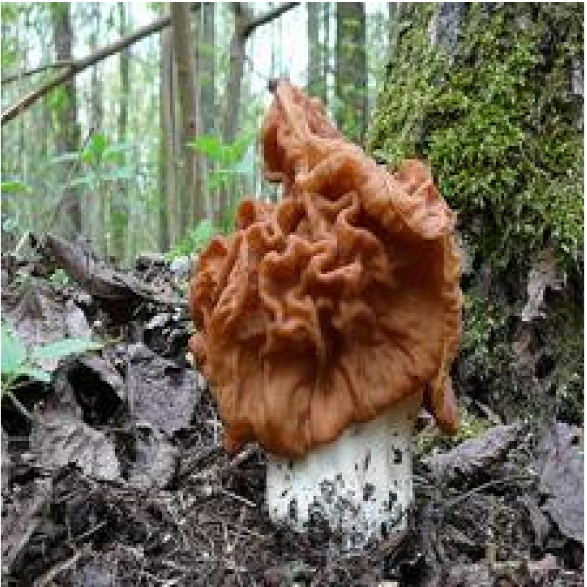}
        \end{subfigure}
        \begin{subfigure}[t]{0.23\linewidth}
        \caption*{2 Clusters}
        \includegraphics[width=\textwidth]{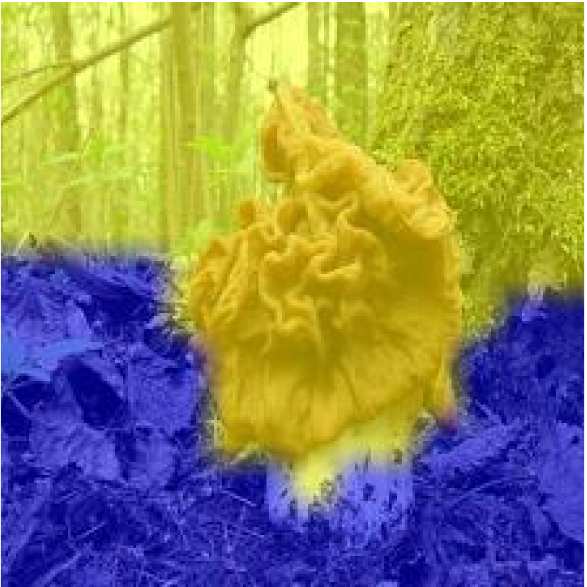}
        \end{subfigure}
        \begin{subfigure}[t]{0.23\linewidth}
        \caption*{3 Clusters}
        \includegraphics[width=\textwidth]{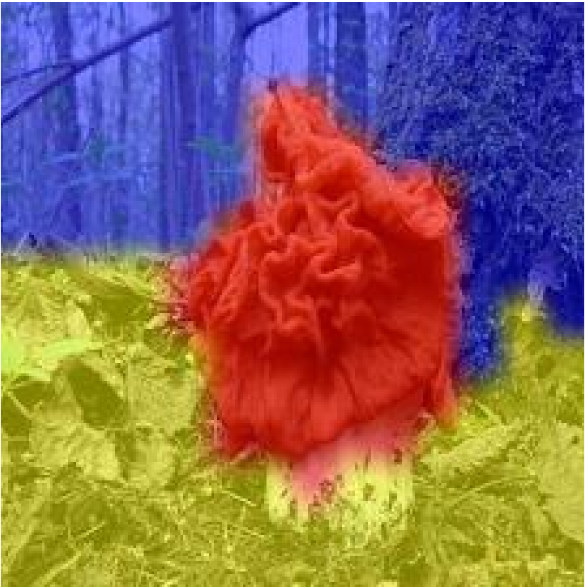}
        \end{subfigure}
        \begin{subfigure}[t]{0.23\linewidth}
        \caption*{4 Clusters}
        \includegraphics[width=\textwidth]{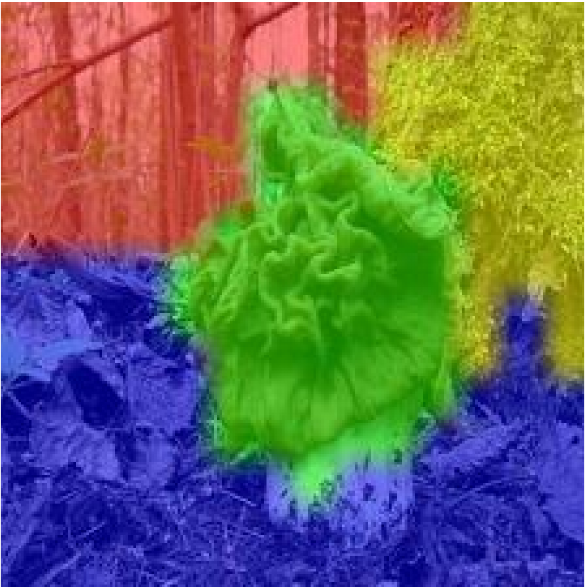}\end{subfigure}
        
        \vspace{0.1cm}
        \end{subfigure}
    \begin{subfigure}[t]{\linewidth}
        \centering
        \includegraphics[width=0.23\textwidth]{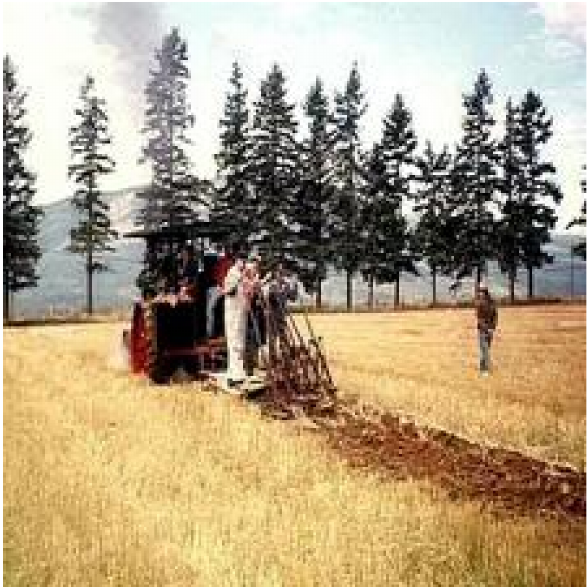}
        \includegraphics[width=0.23\textwidth]{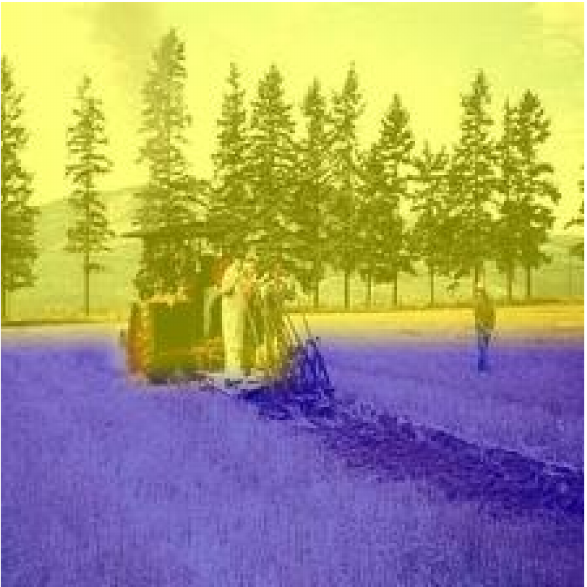}
        \includegraphics[width=0.23\textwidth]{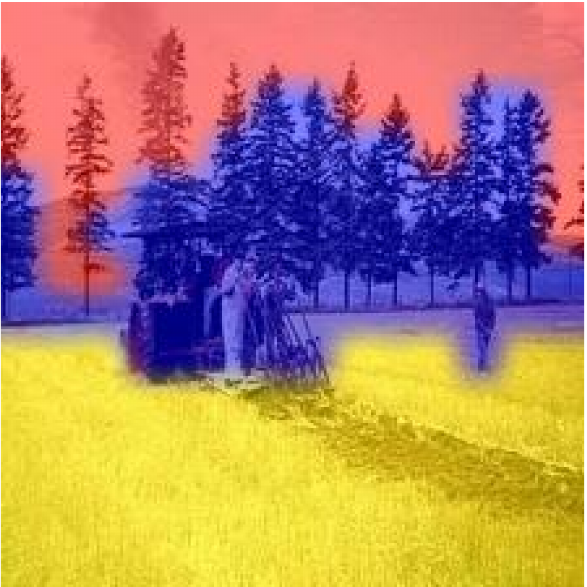}
        \includegraphics[width=0.23\textwidth]{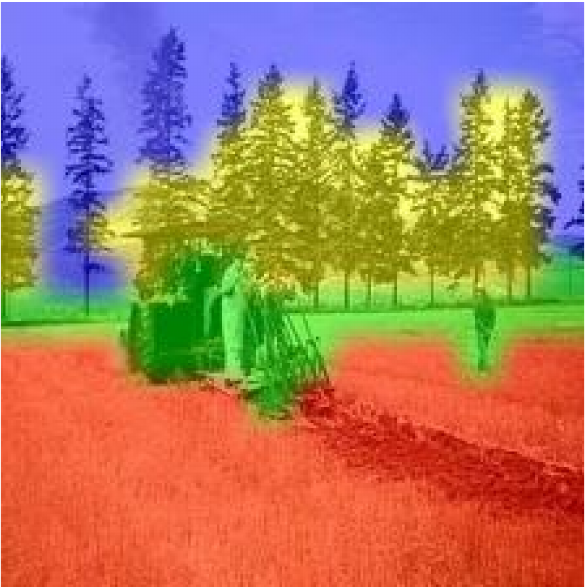}
                
    \vspace{0.1cm}
    \end{subfigure}

    \begin{subfigure}[t]{\linewidth}
        \centering
        \includegraphics[width=0.23\textwidth]{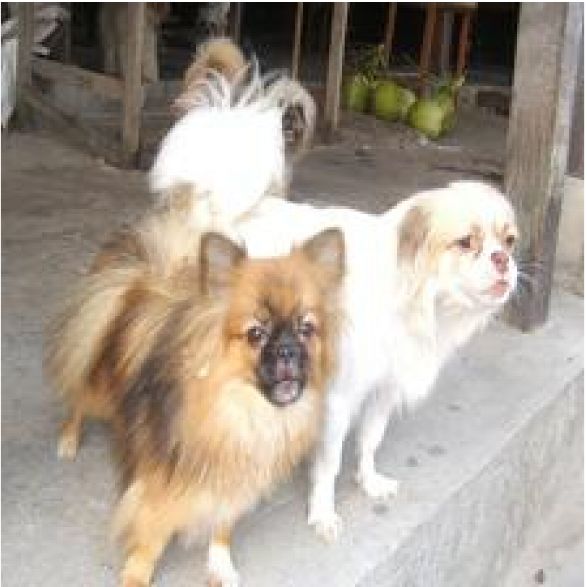}
        \includegraphics[width=0.23\textwidth]{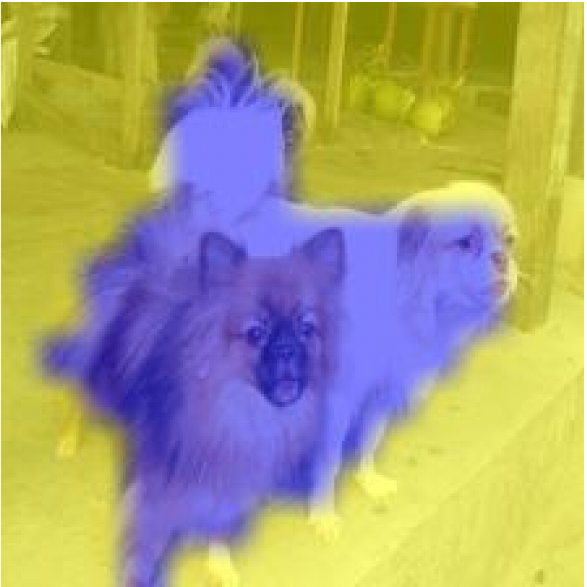}
        \includegraphics[width=0.23\textwidth]{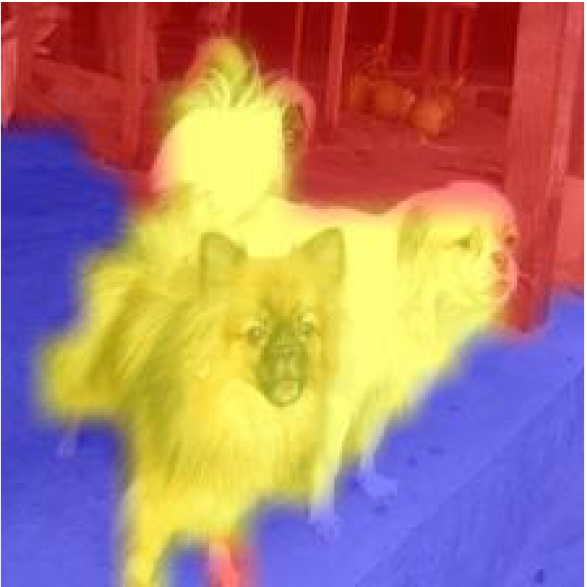}
        \includegraphics[width=0.23\textwidth]{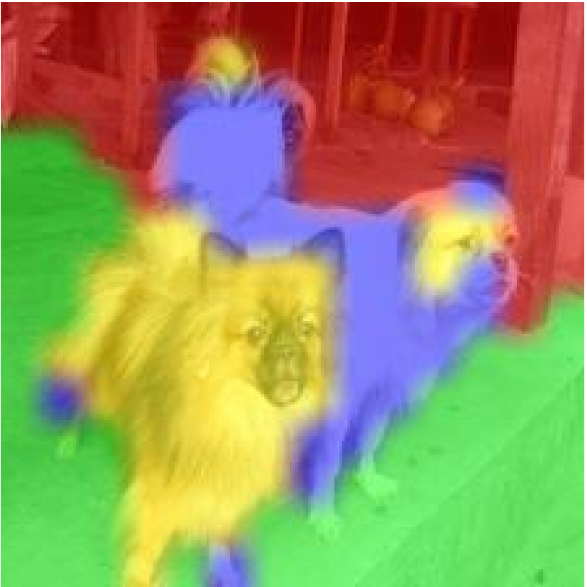}
                
    \vspace{0.1cm}
    \end{subfigure}
    \begin{subfigure}[t]{\linewidth}
        \centering
        \includegraphics[width=0.23\textwidth]{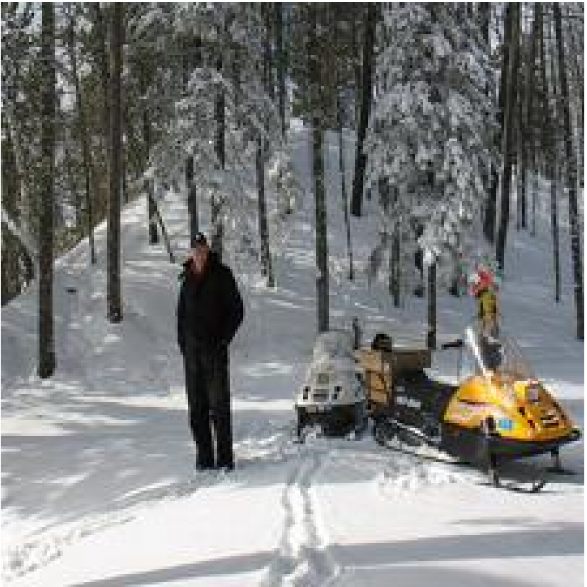}
        \includegraphics[width=0.23\textwidth]{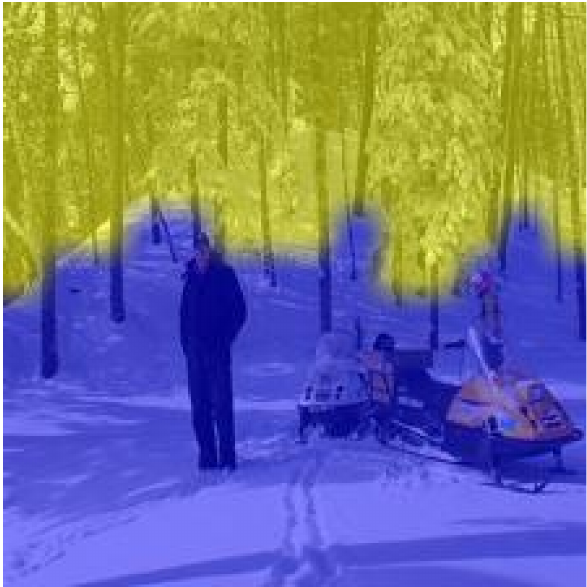}
        \includegraphics[width=0.23\textwidth]{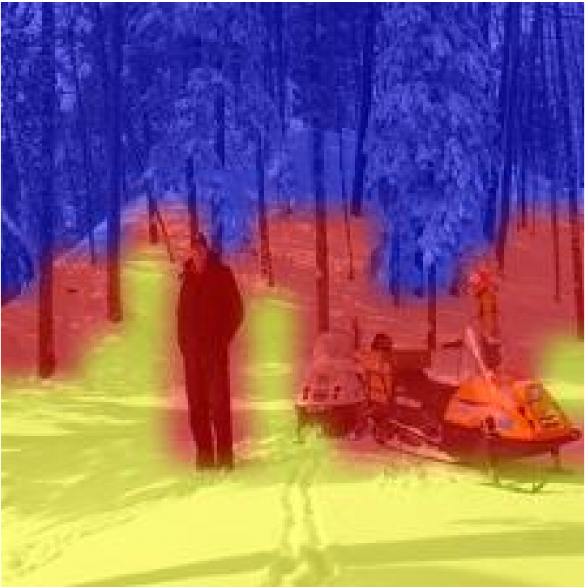}
        \includegraphics[width=0.23\textwidth]{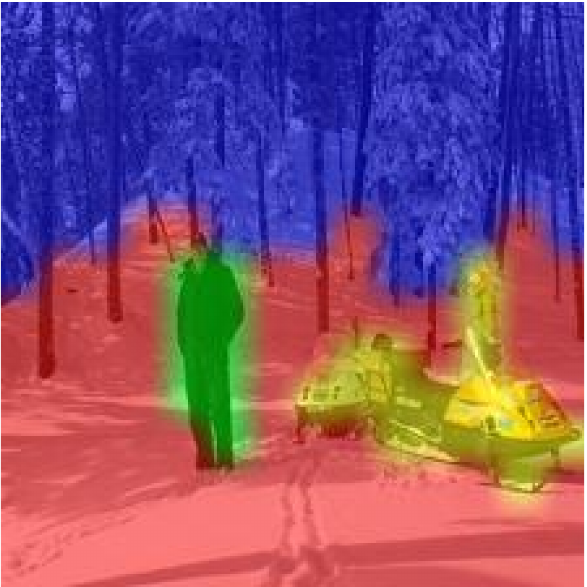}
                
    \vspace{0.1cm}
    \end{subfigure}
    \begin{subfigure}[t]{\linewidth}
        \centering
        \includegraphics[width=0.235\textwidth]{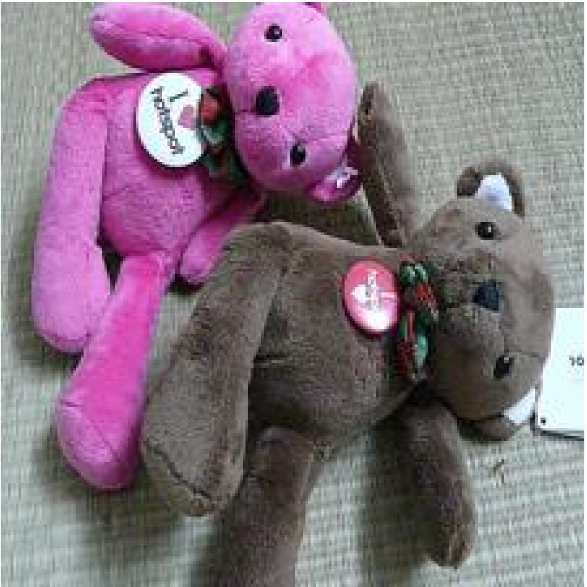}
        \includegraphics[width=0.235\textwidth]{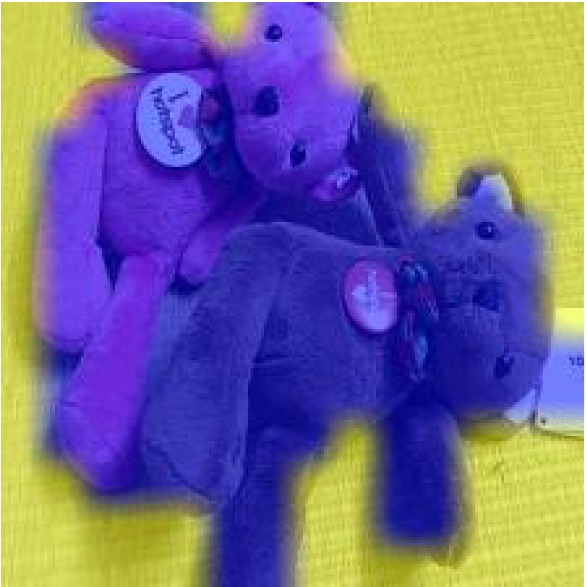}
        \includegraphics[width=0.235\textwidth]{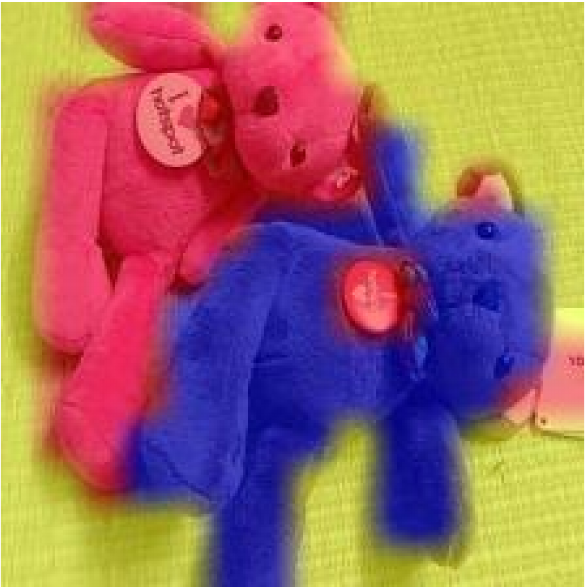}
        \includegraphics[width=0.235\textwidth]{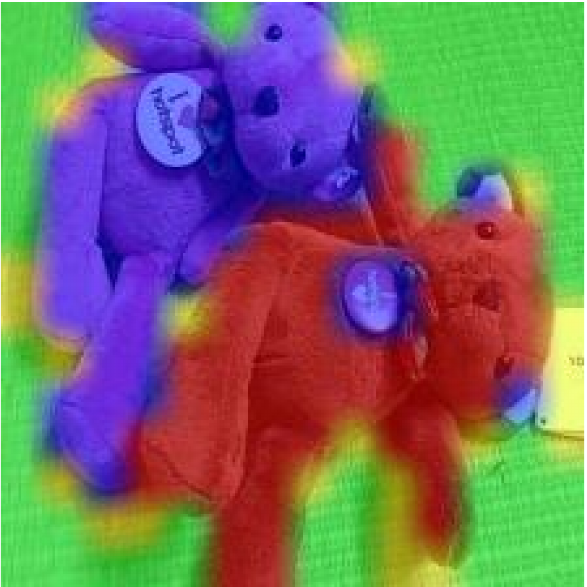}
    \end{subfigure}
    \caption{MC-SSL0.0 has basic knowledge of objects as shown by the self-learnt grouping of data-tokens (the data token on the same object has similar representation) without any labels. Notice how the concepts are refined when asking for more concepts to be discovered. For example, asking for 2 concepts gives forest floor and vegetation, 3 concepts gives forest floor, forest, and mushroom, and 4 concepts gives additional moss on tree. Still a long way to go for MC-SSL as shown by spread out representation of the third concept. This demands for training on bigger datasets and design of more advanced MC-SSL methods. MC-SSL0.0 is pretrained on only 10\% of ImageNet. 
    }
    \label{fig:vis}
    \vspace{-0.35cm}
\end{figure}

\section{Introduction}
\label{sec:intro}

Recent advances in self-supervised learning~\cite{caron2020unsupervised,grill2020bootstrap,chen2020big,atito2021sit,bao2021beit,caron2021emerging} have shown great promise for downstream applications, particularly for image classification datasets with labels for one dominant concept per image (also known as multi-class datasets, \textit{e.g.}  ImageNet~\cite{krizhevsky2012imagenet}). 
These state-of-the-art approaches train the system to extract and associate image features with a single dominant concept, but ignore the intrinsic multi-label nature of natural images that depict more than one object/concept. 
A study by Stock and Cisse~\cite{stock2018convnets} provided the empirical evidence, proving that the remaining error in the ImageNet dataset is predominantly due to the single-label annotation. 
Indeed every pixel in an image is part of some semantic concept with no such thing as background. However, collecting an exhaustive list of labels for every single object/concept is not scalable and requires a significant human effort, making large scale supervised training infeasible for multi-label datasets.  
In view of this, it is pertinent to ask what the best strategy is to move forward in the presence of these deficiencies. We believe that multi-concept self-supervised learning (i.e. ability to represent each concept/object in an image without using labels) is the principled way forward. 

The main aim of this study is to make a step towards building a self-supervised framework capable of learning representations for all the objects in an image. 
Once the multiple-concepts are extracted without supervision, the expensive multi-label 
annotated datasets will be needed only  for evaluation and testing. 
Accordingly, we introduce MC-SSL0.0, a smart framework for self-supervised learning designed to model the meaningful, important, semantic information present in every patch in the image.

MC-SSL0.0 is built on Group Masked Model Learning (GMML) introduced in \cite{atito2021sit}. In MC-SSL0.0, the network is trained with two objectives: i) to reconstruct the raw pixel values of the GMML-based manipulated data-tokens, and, crucially, ii) learning patch level concepts/classes for individual data-tokens. 
MC-SSL0.0 requires the network to learn the knowledge about an object/concept (that is properties such as colour, texture and structure, as well as context) in order to reconstruct, as well as to recover the distorted data-tokens by using available information in unmasked data tokens on the object and its surroundings. This encourages all the data-tokens on an object to have similar representation to each other and incorporate local context in transformers (c.f Figure~\ref{fig:vis}). The ultimate role of the auxiliary but complementary task of learning a patch classifier is to assign a pseudo-semantic label to a group of context aware data tokens covering an object. Our conjecture is that learning pseudo labels for patches encourages data tokens on similar objects within an image and between images to belong to the same pseudo class promoting intra and inter image concept consistency (see Appendix \ref{sec:append}). 
An object consists of group of related data token sharing a common structure captured by a limited set of token level pseudo labels. 
This contextual discovery/learning of objects across a collection of images can conform to human semantic labels, e.g., 
by weak supervision corresponding to cluster of common objects. 

The main contribution of the work is introducing a novel SSL framework which is not compromised by the unrealistic premise that an image contains just a single object, and even if there are other, non dominant objects, they do not impact on the extracted representation. The proposed framework makes a better use of the information conveyed by  all the objects/concepts present in the image. MC-SSL0.0 is a step towards getting representations for each of the object/concept in an image, without the need for any labels. Besides, MC-SSL0.0 enables the training of data hungry transformers from scratch with only a few thousand images. The possibility of training from scratch on limited data with high accuracy will have a significant impact on small AI research groups, companies and application domains which are mainly relying on pretrained models. 
Additionally, we show that, although MC-SSL0.0 is unaware of semantic concepts present in an image, as evident from Figure~\ref{fig:vis}, the self-learnt grouping of data-tokens corresponds to semantic concepts without using any labels. The impact of the proposed innovation is that MC-SSL0.0 outperforms state-of-the-art SSL methods by a large margin in multi-label classification tasks, and achieves competitive results in multi-class tasks. Lastly, MC-SSL0.0 based self-supervised pretraining outperforms supervised pretraining for downstream tasks.

\begin{figure*}[t]
    \centering
    \includegraphics[width= \linewidth]{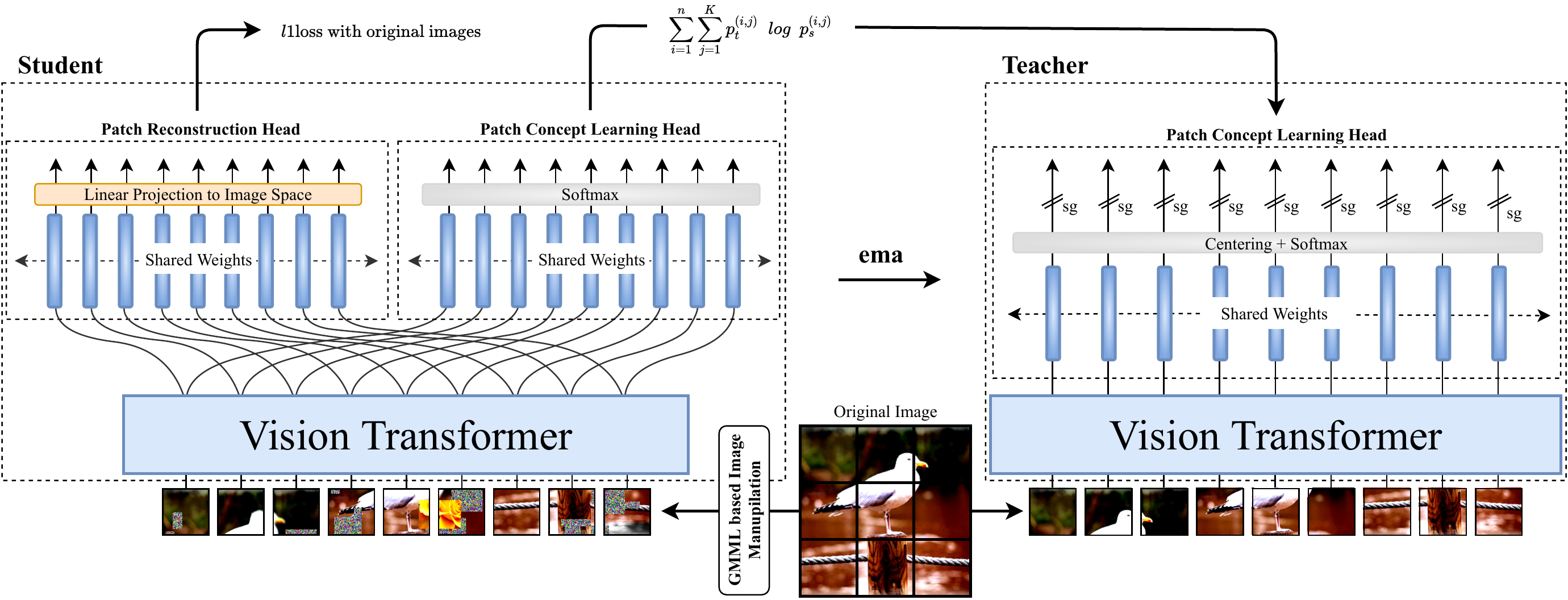}
    \caption{Proposed MC-SSL0.0 framework; A step towards Self-supervised Transformers for Multi-Concept Learning. }
    \label{fig:transformer}
\end{figure*}


\section{Related Work}
\label{sec:RelatedWorks}

Self-supervised methods can roughly be categorised to generative and discriminative approaches. Generative approaches~\cite{doersch2015unsupervised,dumoulin2016adversarially,NEURIPS2019_18cdf49e} learn to model the distribution of the data. However, data modelling generally is computationally expensive and may not be necessary for representation learning in all scenarios. On the other hand, discriminative approaches, typically implemented in a contrastive learning framework~\cite{caron2018deep,hjelm2018learning,chen2020simple,NEURIPS2020_29539ed9,grill2020bootstrap,caron2020unsupervised,caron2021emerging,zbontar2021barlow,bardes2021vicreg,chen2021empirical} or using pre-text tasks~\cite{larsson2017colorization,gidaris2018unsupervised,jenni2018self,atito2021sit,bao2021beit,Kaiming2021mae}, demonstrate the ability to obtain better generalised representations with modest computational requirements.

A large body of work on contrastive learning train the model to discriminate between images considering each of them as a different class. Such approaches require either large batches~\cite{chen2020simple,chen2021empirical} or memory banks~\cite{caron2018deep,he2020momentum} to perform well which is often computationally expensive.

Another recent line of work has shown that it is possible to learn feature representations that are invariant to different augmented views of the same image without requiring negative samples~\cite{caron2020unsupervised,chen2021exploring,grill2020bootstrap,zbontar2021barlow,bardes2021vicreg,caron2021emerging}. 
Such approaches are prune to trivial embeddings. Grill~\etal~\cite{grill2020bootstrap} prevent collapse by employing a momentum encoder with an asymmetric predictor. Barlow Twins~\cite{zbontar2021barlow} and VICREG~\cite{bardes2021vicreg} employed the covariance criterion's where in Barlow Twins for example, the model is trained to obtain an identity cross-correlation matrix between the outputs of two identical networks fed with the augmented versions of a given image. 
On the other hand, Caron~\etal~\cite{caron2021emerging} proposed centring trick by preventing one dimension to dominate.

Despite the impressive results achieved by contrastive learning methods, they often encourage modelling of one dominant class per image and/or disregard the learning of contextual representations, for which alternative pretext tasks might be better suited, including inpainting patches~\cite{pathak2016context}, colourisation~\cite{zhang2016colorful,larsson2016learning,larsson2017colorization}, relative patch location~\cite{doersch2015unsupervised}, solving jigsaw puzzles~\cite{noroozi2016unsupervised,kim2018learning}, cross-channel prediction~\cite{zhang2017split}, predicting noise~\cite{bojanowski2017unsupervised}, predicting image rotations~\cite{gidaris2018unsupervised}, spotting  artefacts~\cite{jenni2018self}, etc.

One of the earliest pioneering work combining self-supervised learning and vision transformers is SiT~\cite{atito2021sit}. SiT introduced two simple but key concepts for self-supervised learning of vision transformers 1) Group Masked Model Learning (GMML), 2) Use of masked auto-encoder. 
The idea of GMML is to learn a neural network by transforming a group of connected patches having significant semantic meaning and recover them by using visible contextual data.  
Beit~\cite{bao2021beit} employed the GMML but instead of recovering the pixel values, the network is trained to predict the visual tokens that are corresponding to the masked patches using the publicly available image tokenizer described in \cite{ramesh2021zero}. He~\etal~\cite{Kaiming2021mae} verified that GMML and masked autoencoder provide a strong pretext task in the case of full ImageNet leading to state-of-the-art performance of multiple downstream tasks. 
Different from \cite{bao2021beit} which employs a pretrained image tokenizer, in this work, we propose a novel SSL framework based on the idea of GMML for masked autoencoder and learning of pseudo-concepts for data-tokens based on knowledge distillation.  

\section{Methodology}
\label{sec:method}
In this section, we introduce the self-supervised image transformer which is a step towards multi-concept self-supervised learning (MC-SSL). Our proposed framework is based on the GMML and self-learning of data token conceptualisation with the incorporation of knowledge distillation~\cite{hinton2015distilling}. 
In Knowledge distillation a student network $\textbf{s}^{\theta}(.)$ is trained to match the output of a given teacher network $\textbf{t}^{\phi}(.)$, where $\theta$ and $\phi$ are the parameters of the student and the teacher networks, respectively. In this work, we employ the same network architecture for both the student and the teacher (i.e. the teacher is a momentum encoder~\cite{he2020momentum} of the student), where the parameters of the teacher network are updated from the past iterations of the student network using exponential moving average of the student weights with the following update rule: $\phi = \lambda \phi + (1 - \lambda) \theta$.

The network architecture of the student network comprises three components: A vision transformer backbone $s_b(.)$, followed by two projection heads attached to the output of the transformer. One projection head is for patch reconstruction $s_{pr}(.)$, and the second projection head is for patch classification $s_{pc}(.)$. The main architecture of our proposed approach is shown in Figure \ref{fig:transformer}.

\subsection{Self-Supervised Vision Transformer}
Similar to \cite{dosovitskiy2020image}, we use vision transformer, which receives as input a feature map from the output of a convolutional block/layer. The convolutional block takes an input image $\mathbf{x} \in \mathbb{R}^{H \times W \times C}$ and converts it to feature maps of size $\mathbf{{x^f}} \in \mathbb{R}^{\sqrt{n} \times \sqrt{n} \times D}$, where, $H$, $W$, and $C$ are height, width and channels of the input image, $n$ is the total number of spatial locations in the feature maps and $D$ is the number of feature map channels. Each spatial location in the input feature maps is considered as an input data token to the transformer,  yielding  a total of $n$ tokens. Both the input and output of the transformer have the same dimensions $\mathbf{x^f, y} \in \mathbb{R}^{n \times D}$.

\subsection{Self-Supervised Tasks}

For SSL, we adopted the GMML as the key component of MC-SSL0.0. 
In NLP, a single data-token can represent a semantic concept, hence, an effective pretext task is to randomly mask the data-tokens and learn to recover them. However, this naive masking might not be particularly effective in CV as a single data token may not represent a semantic concept. The idea of GMML is to transform a group of connected patches representing a significant part of an object having semantic meaning and recover them by learning a model. 
To consolidate the semantically related data-tokens forming an object, GMML is performed by applying several transformations to locally connected data-token/patches of the image, including random drop (i.e. replacing random connected patches with noise) and random replace (i.e. replacing random connected patches with patches from another image), colour distortions, recolouring, etc. Note that unlike NLP, GMML in computer vision (CV) has an added feature, where partial transformation of data-token is possible. The transformer is then trained to recover the values of the corrupted/missing data-token to learn better semantic representations of the input images. 

The main strength of MC-SSL0.0 is that the group of missing/corrupted tokens can only be recovered if the transformer has learnt the context of the object from visible patches. Another strength of MC-SSL0.0 is its desirable learning behaviour. Unlike, other SSL losses, MC-SSL0.0 performance saturates slower, leading to continuous improvement in performance. This could be because a large portion of image is masked, providing heavy augmentation for training data. Nevertheless, the performance gain becomes marginal after a few hundred epochs. 
It should be noted that these transformed tokens can either be  on the dominant (so called foreground) object or on the other (so called background) objects, and recovering these tokens is equally valid for both scenarios. 
Our thesis is that there is no such thing as background in natural images. Each patch/pixel in the image represents some concept with visual semantic meaning. 
The intuition is that by modelling all semantic concepts, the network will generalise better for unseen tasks, whether they are related to an object, a distributed object, dense prediction task like detection and segmentation or to the whole visual signal. 

We leverage the strength of the transformers and incorporate it with GMML to train MC-SSL0.0 with two different objectives: Patch reconstruction (Section~\ref{sec:pr}), where the network is trained to reconstruct the image corrupted by the GMML-based manipulation and patch concept classification (Section~\ref{sec:pc}) by training the network to learn data-token level concepts/classes for individual data-tokens.

\subsubsection{Patch Reconstruction}
\label{sec:pr}

For image reconstruction, we propose to use the transformer as a group masked autoencoder, \textit{i.e.}, visual transformer autoencoder with GMML. By analogy to auto-encoders, our network is trained to reconstruct the input image through the output tokens of the transformer. 
The GMML-based manipulated images $\mathbf{\bar{x}} \in \mathbb{R}^{H \times W \times C}$ are fed to the student backbone and the output tokens of the student transformer are fed to the patch reconstruction projection head $s_{pr}(.)$ to obtain $\mathbf{x_r} \coloneqq s_{pr}(s_b(\mathbf{\bar{x}}))$, i.e. the reconstructed image. The $s_{pr}(.)$ projection head consists of 3 fully connected layers; first two with $2048$ neurons and GeLU~\cite{hendrycks2016gaussian} non-linearity each, and the last bottleneck layer with $256$ neurons, followed by a transposed convolution to return back to the image space. 

The objective of the image reconstruction is to restore the original image from the corrupted image. For this task, we use the $\ell1$-loss between the original and the reconstructed image as shown in Equation \ref{eq:l1-pixel}. Although, $\ell2$-loss generally converges faster than $\ell1$-loss, $\ell2$-loss is prone to over-smooth the edges in the restored image~\cite{zhao2016loss}. Therefore, $\ell1$-loss is commonly used for image-to-image processing rather than the $\ell2$-loss.

\begin{equation}
\label{eq:l1-pixel}
\mathcal{L}_{\rm recons}(\mathbf{W}) = ||\mathbf{x} - \mathbf{x_r}||
\end{equation}
where $||.||$ is the $\ell1$ norm, 
and $\mathbf{W}$ denotes the parameters of the transformer to be learned during training. 

\subsubsection{Patch Concept Learning}
\label{sec:pc}

The idea of concept learning using SSL is first introduced in DINO~\cite{caron2021emerging}. DINO is providing a pseudo label for the student network by setting a low temperature in the softmax activation of the teacher network. This low temperature sharpens the probability vector leading to one class getting significantly higher probability than the others. Hence, DINO is focusing on learning one dominant concept in the image.
Following our hypothesis that modelling only the dominant class in an image can lead to sub-optimal representation learning, we investigate the role of learning the appropriate concept for each of the data token. This is a step towards extracting a representation corresponding to each concept in an image.  
In a recent work, Beit~\cite{bao2021beit} assigned fixed classes to each of the patches using a pre-trained image tokenizer~\cite{ramesh2021zero}. Then they used BERT/SiT like framework to predict the classes corresponding to a masked data token. 
Unlike Beit, we employ a momentum encoder (teacher) to generate pseudo labels for the visual tokens, and force the student model to make predictions consistent with the teacher model. Patch concept learning gives the flexibility to adapt to the visual concepts present in images rather, than using a pretrained fixed tokenizer. This is an import ingredient in learning the semantic representation of each object, which will be described at the end of the section. 

In order to generate pseudo labels for the visual tokens, the training images are fed to the backbone of the teacher network, and the outputs of the data tokens are then fed to a patch classification projection head to obtain $z_t \coloneqq t_{pc}(t_b(\textbf{x})) \in \mathbb{R}^{n \times K}$, where $K$ represents the number of classes of visual tokens. Similar to DINO, the patch classification projection head consists of 3 fully connected layers; first two with $2048$ neurons and GeLU non-linearity each, and the last bottleneck layer with $256$ neurons. The output of the bottleneck layer is $l2$ normalised and directly connected to a weight-normalised fully connected classification layer with $K$ neurons. 
For each training sample, the GMML-based manipulated images are passed to the student network to obtain $z_s = s_{pc}(s_b(\mathbf{\bar{x}}))\in \mathbb{R}^{n \times K}$. The task is to match $pc_s$ to $pc_t$ employing the Kullback-Leibler divergence (KL) between the outputs of the teacher and student networks. 

Training the student to match the teacher output can easily lead to a trivial constant (i.e. collapsed) embeddings. To avoid the model collapse, we adopted the centring and sharpening of the momentum teacher outputs introduced in~\cite{caron2021emerging}. Centring encourages the output to follow a uniform distribution, while the sharpening has the opposite effect. Applying both operations balances their effects, which is sufficient to avoid a collapse in the presence of a momentum teacher. The centre $c$ is updated using an exponential moving average over the teacher output. The sharpening is obtained by using a low value for the temperature $\tau_t$ in the teacher softmax normalisation. The output probability distributions $p_t$ and $p_s$ from the teacher and the student networks over $n$ patches and $K$ dimensions are obtained as follows:
\begin{equation}
    p_s^{(i, j)} = \frac{\exp({z_s^{(i, j)}}/{\tau_s})}{\sum_{k=1}^K \exp({z_s^{(i, k)}}/{\tau_s})}
\end{equation}
\begin{equation}
    p_t^{(i, j)} = \frac{\exp({z_t^{(i, j)}}/{\tau_t})}{\sum_{k=1}^K \exp({z_t^{(i, k)}}/{\tau_t})} 
\end{equation}
where, $z_s$ and $z_t$ are the class logits for the student and the teacher, $p_s(i, .)$ and $p_t(i, .)$ are the output probabilities by the student and teacher, corresponding to the $i^\text{th}$ token, and $\tau_t$ and $\tau_s$ are the temperature parameters for the teacher and the student, respectively.

\begin{equation}
    \mathcal{L}_{\rm classify}(\mathbf{W}) = \frac{1}{n \times K}  \sum_{i=1}^n \sum_{j=1}^K  p_t^{(i, j)} \log p_s^{(i, j)}
\end{equation}

\subsection{Putting together the MC-SSL0.0 framework} 
The reconstruction of the grouped masked tokens and learning of patch level concepts with MC-SSL0.0 gives us a mechanism to adaptively learn multiple concepts in each image. The hypothesis of MC-SSL0.0 is that a network will be able to recover semantically transformed data-tokens from the rest of the semantic clues about the object and context only if it learns the ``semantics'' of the objects in the image. Once transformer autoencoder is aware of a semantic concept, the role of the auxiliary patch concept learning task is to encourage a shared/common label corresponding to all semantically related data-tokens. 
This consolidated information from all the data-tokens about an object (distributed concept) in an image can be assigned a name which humans are using in their daily life. 
As a data-token represents a small portion of image/object and is likely to have less overall variability, we do not need to learn a large number of classes/concepts for patches, i.e., $K$ can be small. 
In contrast, a complex object needs to consolidate information from multiple data-tokens, which constitute the object. Hence, even with a small/limited number of local concepts for each data-token, the possible representation space for objects is huge. For example, if the object consists of only four data-tokens, and a patch concept learning space is only 1024, the number of possible configurations of local concepts will be $1024^4 > 1.9 trillions$. However, due to the local stationarity of natural visual signals the actual combination will be far less. 

Transformer autoencoder and auxiliary task of patch concept learning\footnote{We note that in parallel with us, iBot~\cite{Zhou2021iBOTIB} used the idea of patch concept learning. However, iBot focuses on learning a dominant concept via the classification token with DINO loss and hence, models the dominant class, which we believe is a limitation of existing SSL methods.} in MC-SSL0.0 provide a superior way to use the information present in an image. More importantly, it give us a mechanism to model all the concept/object present in an image. 
Some self learnt concepts in each images are shown in Figure~\ref{fig:vis}. Specifically, we obtain $y_t \coloneqq t_b(\textbf{x})\in \mathbb{R}^{n \times D}$ which are the output features corresponding to the data-tokens of the input image. The output features are then clustered, employing simple k-means~\cite{arthur2006k}), into 2, 3, and 4 clusters, where each colour represents a different cluster. Note that the model is trained without any sort of supervision, yet, MC-SSL0.0 demonstrates the ability to differentiate between concepts. As shown in Section~\ref{sec:exp} this superior way of utilising the information enables us to obtain remarkable results with extremely limited resources, limiting us to use only small models and 10\% of ImageNet. 

\subsection{Properties of MC-SSL0.0}

Some of the desirable properties of MC-SSL0.0 include:
\begin{enumerate}
    \item Training Transformers on tiny datasets: Transformers trained by supervised losses can attend to all the data-token coming from an image empowering them to model global information. However, the side effect is that they need a lot more data to model local context (stationary of visual signal in local neighbourhood) 
    MC-SSL0.0 overcome this limitation by masking group of semantically related data-token. This masking encourages the network to learn from semantic information present in the locality of group of masked tokens to recover the missing information hence, modelling local contextual information. Therefore, using MC-SSL0.0 based pretraining it is possible to train the transformers on small datasets with high accuracy. In fact we have validated this by training vision transformer from scratch for a high accuracy using MC-SSL0.0 as pretext task. 
    \item The MC-SSL0.0 framework is aware of the notion of concepts in the sense that the heavily masked information can be reconstructed with respect to shape and texture from the available information from the visible data tokens on the same concept/object and surrounding concepts. This is also evident from the ability of MC-SSL0.0 to self-cluster semantically related data-tokens from an object in the image without using any labels for training as demonstrated in Figure~\ref{fig:vis}. 
    \item MC-SSL0.0 based pretraining outperforms supervised pretraining for multiple downstream tasks of multi-label and multi-class datasets given the same amount of pretraining data. Moreover, MC-SSL0.0 pretraining also outperforms state-of-the-art SSL pretraining. This is due to the fact that MC-SSL0.0 makes better use of rich semantic information present in all the data-tokens relating to each concept/class/object in an image. 
    \item Big batch size is a standard requirement in many of the constrastive learning based SSL, making them unusable when relying on a modest GPU resource. MC-SSL0.0 does not suffered from this problem and outperforms the state-of-the-art for small batch sizes as well. This strength is coming from the nature of the pretext task which does not involve negative classes. 
    \item Proposed framework is generic and is applicable to other application domains of AI, e.g., sound, medical image analysis, etc. We left this for future study. 
\end{enumerate}

\section{Experimental Results}
\label{sec:exp}

\noindent{Disclaimer/Preliminary: We are demonstrating most of the results on 10\% of ImageNet dataset due to the fact that the research is coming from a small group with 1.6 GPU per researcher on average.
We hope that the inability to perform large scale experiments will not be used to penalise less resourced groups, but instead, the focus will be on the scientific contribution, and its demonstrated merits, as well as its potential impact on large scale deep learning problems.} 

We follow the common evaluation protocol to demonstrate the generalisation of the learnt features of the proposed MC-SSL0.0 approach by pretraining the model in unsupervised fashion, followed by finetuning on a downstream task. We conduct several experiments on multi-label and multi-class image classification downstream tasks. 

To the best of our knowledge, most of the SSL SOTA methods are validated on multi-class downstream tasks and only few SSL methods \cite{caron2018deep,feng2019self} involved multi-label classification as a downstream task where simple CNN architectures, like AlexNet \cite{krizhevsky2012imagenet}, were employed.

Therefore, to have a fair comparison with SOTA, we also pre-trained the well respected SSL algorithm DINO \cite{caron2021emerging} on 10\% of the ImageNet using their official code with the suggested default parameters and compare the performance with MC-SSL0.0 in both of the multi-label and the multi-class downstream tasks. Due to the limited resources, we only compare to DINO with the premise that DINO reflects SOTA accuracy reasonably well.

Based on the observation of Caron \etal~\cite{caron2021emerging}, multi-crop training in DINO is an important component to obtain good features. Although it is not required in our proposed approach, we also adopted the multi-crop strategy to match the number of network updates and have unbiased comparison with DINO.


In Section \ref{sec:impDet}, we provide the implementation details to
train the proposed MC-SSL0.0 model in a self-supervised fashion. Next, we explain the datasets, the evaluation metrics, and the results for the multi-label and multi-class classification tasks in Section \ref{sec:multilabel} and \ref{sec:multiclass}, respectively. 
Furthermore, we conduct several ablation studies to investigate the effect
of the individual components of the proposed approach in the Appendix \ref{sec:append}.

\subsection{Implementation Details}
\label{sec:impDet}

The backbone architecture of MC-SSL0.0 employs a small variant of vision transformer (ViT-S/16)~\cite{dosovitskiy2020image}. Our augmentation policy follows the SiT~\cite{atito2021sit} policy, which includes random resized cropping, horizontal flipping, followed by randomly applying color jittering, converting to grey-scale,, blurring, random drop, and random replace.

For the optimisation of the self-supervised training, the model is trained using the Adam optimiser~\cite{Loshchilov2017FixingWD} with a batch size of $64$ images per GPU, and momentum of $0.9$. The weight decay follows a cosine schedule~\cite{loshchilov2016sgdr} from $0.04$ to $0.4$, and the base learning rate is $5e^{-4}$. All the models are trained for $500$. Similar to DINO \cite{caron2021emerging}, the sharpening parameters of the teacher and the student are set to $\tau_t=0.04$ and $\tau_s=0.1$. The teacher is updated using exponential moving average on the student weights with $\lambda$ following a cosine schedule from $0.996$ to $1$ during training. 

Regarding the objective function, we combined the patch classification loss and the patch reconstruction loss using simple averaging. We believe further improvements can be gained by optimising the weighted sum of the losses or by incorporating  the  uncertainty  weighting  approach  proposed by Kendall \etal~\cite{kendall2018multi}.

Last, for the downstream tasks, the projection heads are discarded and finetuning is performed employing the backbone of the pretrained teacher network $t_b(.)$.

\subsection{Multi-Label Classification}
\label{sec:multilabel}

\noindent
\textbf{Experimental Setup. }
The models are pre-trained in unsupervised fashion using the ViT-S/16 backbone on 10\% of the ImageNet-1K dataset, with an input size $224\times224$. For the finetuning step on the multi-label downstream tasks, the projection heads are replaced with a fully connected layer with $2048$ neurons with the GeLU activation function, followed by an output layer with $c$ nodes, corresponding to the number of classes in the downstream tasks, followed by a Sigmoid activation function. 
For the optimisation, we mostly rely on the vision transformer developer's default hyper-parameters \etal~\cite{touvron2020training} due to the limited resources. We believe that further improvements can be obtained by tuning the hyper-parameters. For the input size and data augmentation, we follow the most common settings \cite{chen2019learning,chen2020knowledge,lanchantin2021general}  where the images are resized to $448\times448$ and augmented with Rand-Augment \cite{cubuk2020randaugment}. The test images are centre cropped instead. All the models are trained for $80$ epochs with $48$ batch size employing 2 Nvidia Tesla V100 32GB GPU cards. 
As for the loss function, we adopted the asymmetric loss~\cite{ridnik2021asymmetric} to address the sample imbalance problem. Asymmetric loss is a variant of focal loss with different $\gamma$ for positive and negative values. Given the target $t = [t_1, t_2, \dots, t_c] \in \{0, 1\}$, where $c$ is the number of classes, and the output probabilities $p = [p_1, p_2, \dots, p_c]$, the asymmetric loss for each training sample is calculated as follows:

\begin{equation}
    \mathcal{L} = \frac{1}{c} \sum_{i=l}^c \begin{cases}
      (1 - p_i)^{\gamma+} \log(p_i), & t_i = 1,\\
      (p_i)^{\gamma-}\log(1 - p_i), & t_i = 0
    \end{cases}  
\end{equation}

We employed the default values for $\gamma$  in our experiments, \textit{i.e.}, we set $\gamma+=0$ and $\gamma-=4$.

\begin{table}[t]
\centering
\caption{The results of regular inference on the PASCAL VOC 2007, VG-500, and MS-COCO datasets. The threshold is set to 0.5 to compute the precision, recall and F1 scores (\%). All the models are pre-trained with resolution $224\times224$ and finetuned with resolution $448\times448$.}
\label{tbl:VOC}
\resizebox{\linewidth}{!}{
\begin{tabular}{lccccccc}
\hline
Method & ~~mAP~~ & CP & CR & CF1 & OP & OR & OF1\\ \hline
\multicolumn{8}{c}{PASCAL VOC 2007 dataset}  \\ \hline
\multicolumn{8}{l}{\itshape{\textcolor{darkgray}{From scratch (\textit{i.e.}, random initialization)}}}\\ 
ViT-S/16                 & 30.3 & 21.9 & 55.2 & 31.4 & 27.8 & 66.8 & 39.2 \\ \hline
\multicolumn{8}{l}{\itshape{\textcolor{darkgray}{Selfsupervised pretraining on PASCAL VOC}}}\\ 
MC-SSL0.0 ~~~~~~             & 65.2 & 42.6 & 73.1 & 53.8 & 47.9 & 78.6 & 59.6\\\hline
\multicolumn{8}{l}{\itshape{\textcolor{darkgray}{Selfsupervised pretraining on 10\% of ImageNet-1K}}}\\ 
Dino*                     & 72.8 & 49.7 &  75.3 & 59.9 & 55.8 & 80.2 & 65.8\\ 
MC-SSL0.0                    & 79.3 & 63.6 & 80.7 & 71.2 & 67.9 & 84.8 & 75.4\\\hline
\multicolumn{8}{l}{\itshape{\textcolor{darkgray}{Selfsupervised pretraining on 10\% ImageNet-1K with multi-crop}}}\\ 
$\text{Dino}^\ddagger$    & 80.2 & 56.1 & 80.5 & 66.1 & 62.1 & 84.7 & 71.7\\
$\text{MC-SSL0.0}^\ddagger$  & 81.3 & 59.4 & 80.0 & 68.2 & 65.5 & 84.1 & 73.7\\\hline
\multicolumn{8}{c}{VG-500 dataset}  \\ \hline
\multicolumn{8}{l}{\itshape{\textcolor{darkgray}{From scratch (\textit{i.e.}, random initialization)}}}\\ 
ViT-S/16* & 22.1 & 16.4 & 37.8 & 22.9 & 26.4 & 57.5 & 36.2\\ \hline
\multicolumn{8}{l}{\itshape{\textcolor{darkgray}{Selfsupervised pretraining on VG-500}}}\\
MC-SSL0.0 ~~~~~~& 30.0&21.4&46.4&29.3&30.8&64.2&41.7\\\hline
\multicolumn{8}{l}{\itshape{\textcolor{darkgray}{Selfsupervised pretraining on 10\% of ImageNet-1K}}}\\ 
Dino*  & 24.9 & 18.6 & 40.4 & 25.5 & 27.3 &  56.7 & 36.9\\ 
MC-SSL0.0 & 28.9 & 26.6&44.4&33.3&34.9&60.7&44.3\\\hline
\multicolumn{8}{l}{\itshape{\textcolor{darkgray}{Selfsupervised pretraining on 10\% ImageNet-1K with multi-crop}}}\\ 
$\text{Dino}^\ddagger$  & 27.7 & 19.8 & 45.3 & 27.6 & 28.5 & 62.8 & 39.2\\
$\text{MC-SSL0.0}^\ddagger$ & 29.4 & 20.5 & 46.1 & 28.4 & 30.5 & 64.5 & 41.4\\\hline
\multicolumn{8}{c}{MS-COCO dataset}  \\ \hline
\multicolumn{8}{l}{\itshape{\textcolor{darkgray}{From scratch (\textit{i.e.}, random initialization)}}}\\ 
ViT-S/16*       & 44.7  & 32.7 & 58.7 & 42.0 & 37.1 & 67.9 & 48.0\\ \hline
\multicolumn{8}{l}{\itshape{\textcolor{darkgray}{Selfsupervised pretraining on MS-COCO}}}\\ 
MC-SSL0.0 ~~~~~~& 73.1 & 56.2 & 75.2 & 64.3 & 58.6 & 80.1 & 67.7\\\hline
\multicolumn{8}{l}{\itshape{\textcolor{darkgray}{Selfsupervised pretraining on 10\% of ImageNet-1K}}}\\ 
Dino*  & 63.4 & 50.8 & 66.5 & 57.6 & 54.0 & 73.1 & 62.1\\ 
MC-SSL0.0 & 70.5 & 54.8 & 74.0 & 63.0 & 56.3 &  79.1 & 65.8\\\hline
\multicolumn{8}{l}{\itshape{\textcolor{darkgray}{Selfsupervised pretraining on 10\% ImageNet-1K with multi-crop}}}\\ 
$\text{Dino}^\ddagger$   &  69.0 & 56.0 & 70.1 & 62.2 & 59.4 & 75.4 & 66.5\\
$\text{MC-SSL0.0}^\ddagger$ & 72.7 &  56.8 &  74.1 & 64.3 & 59.6 & 79.0 & 67.9\\\hline
\end{tabular}}
\end{table}

\noindent
\textbf{Evaluation Metrics.}
Following previous works~\cite{lanchantin2021general,liu2021query2label,cheng2021mltr}, beside the mean average precision (mAP), we employ several metrics to better demonstrate the performance of the proposed approach. Under the premise that the predicted label is positive, if the output probability is greater than a threshold (\textit{e.g.}, 0.5), we report the average per-class precision (CP), recall (CR), and F1 score (CF1). We also present the average overall precision (OP), recall (OR), and F1 score (OF1). 

\noindent
\textbf{Datasets.}
To evaluate the proposed self-supervised multi-label approach, we conduct our experimental analysis on several datasets, including PASCAL VOC~\cite{everingham2015pascal}, Visual Genome~\cite{krishna2017visual}, and MS-COCO~\cite{lin2014microsoft}. 
The PASCAL VOC 2007 ~\cite{everingham2015pascal} includes $20$ object categories and it consists of $5,011$ images for training and $4,952$ for evaluation. The Visual Genome dataset~\cite{krishna2017visual} contains $108,077$ images from around $80$K categories. The label distribution of this dataset is quite sparse. Consequently, most of the experiments in the literature are performed on VG-500, introduced in~\cite{chen2019learning}. VG-500 consists of $98,249$ training images and $10,000$ test images including the most $500$ frequent objects.
MS-COCO~\cite{lin2014microsoft} is a large-scale object detection and segmentation dataset. The standard multi-label formulation for MS-COCO includes $80$ objects with an average of $2.9$ labels per image. The dataset consists of $82,081$ images for training and $40,137$ images for evaluation.

\noindent
\textbf{Results.}
In Table \ref{tbl:VOC}, we compare the proposed MC-SSL0.0 with the DINO framework \cite{caron2021emerging} on three different datasets, PASCAL VOC, VG-500, and MS-COCO, respectively. First, we show the results when ViT-S/16 is trained from scratch on the downstream task. Then, we show the performance when the model is pretrained using MC-SSL0.0, and finetuned employing the same downstream dataset. Finally, we report the accuracy when the models are pre-trained with and without multi-crop strategy on 10\% of ImageNet employing MC-SSL0.0 and DINO frameworks. 

From the reported results, it is evident that the training from random initialisation has produced low accuracies as the amount of data available is insufficient to train the transformer. The results significantly improved when the models are pre-trained using MC-SSL0.0 without any external data with $+33$, $+7.9$, and $+28.2$ absolute mAP improvement in PASCAL VOC, VG-500, and MS-COCO datasets, respectively. Further, pretraining with the MC-SSL0.0 framework consistently outperforms DINO, particularly in the absence of multi-crop strategy, where MC-SSL0.0 obtained $+6.5$, $+4.0$, and $+7.1$ absolute mAP improvement in PASCAL VOC, VG-500, and MS-COCO, respectively. 

\subsection{Multi-Class Classification}
\label{sec:multiclass}
We conduct our experimental analysis on standard multi-class classification problems based on object detection and recognition in an unconstrained background, namely, CIFAR-10/CIFAR-100~\cite{krizhevsky2009learning}, Cars~\cite{Carsdataset}, and Flowers~\cite{FlowersDataset}.

For the transfer learning on downstream tasks, the patch concept learning head is replaced by a linear projection head with $c$ nodes corresponding to the number of classes in the downstream task. The input to the linear projection head is the average of the features coming from the data tokens. For the data augmentation, we applied random cropping, random horizontal flipping, MixUp~\cite{zhang2017mixup}, and Auto-Augment~\cite{cubuk2018autoaugment} during training. For optimisation, we follow the same protocol used in Touvron \etal~\cite{touvron2020training}. 

In Table \ref{tbl:ablation_}, we first report the accuracy on the downstream tasks when the models are trained from scratch with random initialisation as a baseline. Then, we reported the results when the same dataset is used for SSL pre-training and finetuning, i.e. without using any external/additional datasets. Finally, we compare MC-SSL0.0 with DINO trained on 10\% of ImageNet with and without multi-crop strategy.

We found that, MC-SSL0.0 enables training the data hungry transformers on stand-alone small datasets with acceptable performance compared to the pre-training with the full ImageNet-1K dataset. Further, MC-SSL0.0 consistently outperforms DINO, with and without multi-crop strategy. In fact, the performance of MC-SSL0.0 without multi-crop is on par with the performance of DINO with multi-crop strategy. Several ablation studies are performed and shown in the Appendix to investigate the effect of the individual components of MC-SSL0.0, the effect of the percentage of the applied corruption to the input images, and the effect of the choice of $K$ during the MC-SSL0.0 pre-training.  

\begin{table}[t]
\centering
\caption{Transfer learning by finetuning pretrained Self-supervised models on different downstream tasks. Self-supervised models are trained using ViT-S/16 model on $10\%$ of ImageNet dataset, followed by finetuning on downstream tasks.}
\label{tbl:ablation_}
\resizebox{\linewidth}{!}{
\begin{tabular}{lcccc}
\hline
& CIFAR10~     & CIFAR100   & ~~Cars~~    & Flowers~  \\ \hline
\multicolumn{5}{l}{\itshape{\textcolor{darkgray}{From scratch (\textit{i.e.}, random initialization)}}}  \\
ViT-S/16        & 91.42 & 70.14 & 10.67 &  54.04\\ \hline
\multicolumn{5}{l}{\itshape{\textcolor{darkgray}{Self-supervised pre-training on the given dataset}}}  \\
$\text{MC-SSL0.0}^\ddagger$& 98.00 & 85.38 & 89.20 & 87.30\\ \hline
\multicolumn{5}{l}{\itshape{\textcolor{darkgray}{Selfsupervised pretraining on 10\% of ImageNet-1K}}}  \\
& \multicolumn{4}{c}{w/o multi-crop} \\ \cline{2-5} 
Dino                    & 97.27 & 81.77 & 82.08  & 92.68\\ 
MC-SSL0.0 & \textbf{97.82}&\textbf{84.98}&\textbf{86.15}&\textbf{95.56}\\\hline
& \multicolumn{4}{c}{with multi-crop} \\ \cline{2-5} 
$\text{Dino}^\ddagger$   & 97.90 & 84.61 & 88.21  & 95.46  \\ 
$\text{MC-SSL0.0}^\ddagger$ & \textbf{98.08} &  \textbf{85.82} & \textbf{90.44}    & \textbf{96.31 } \\ \hline
\end{tabular}
}
\end{table}

\section{Conclusion and Discussion}
\label{sec:conc}

In this paper, we presented a novel self-supervised learning framework (MC-SSL0.0) that enables the extraction of visual representation corresponding to multiple objects in an image without annotations. We demonstrated several advantages of the proposed MC-SSL0.0 framework. First, MC-SSL0.0 can train transformers from scratch with good accuracy on small datasets. Second, MC-SSL0.0 has some notion of semantic information as demonstrated by the ability to reconstruct missing parts of a concept and by self-learnt grouping of data-tokens corresponding to a semantic concept (Figure~\ref{fig:vis}). Third, MC-SSL0.0 outperforms supervised methods for network pretraining. Last, MC-SSL0.0 outperforms the existing state-of-the-art for both multi-class and multi-label downstream tasks, verifying its strengths. 


SSL in CV has made a tremendous progress with self-supervised pretraining, outperforming supervised pretraining. However, there are several open questions, which should be addressed in the future development of SSL. We only pose a few of them for brevity. 
a) Is the kNN style evaluation of SSL methods right? 
b) What should be the preferred choice to evaluate linear probing and downstream applications?
c) Will more suitable evaluation protocols encourage the community to build SSL algorithms for multi-concept representation learning (i.e. algorithms capable to represent each concept/object in an image without using labels)?  
d) Is it possible to build a representation for each of the concept in an image without any label?

The current kNN and linear evaluation of SSL methods on multi-class datasets, like ImageNet, is biasing the SSL research towards modelling the dominant object in image leading to sub-optimal use of information present in the image. 
We argue that multi-label dataset like Pascal VOC, Visual Genome, MS COCO and dense prediction datasets are more suitable for evaluating the generalisation of SSL methods. The multi-label datasets are suitable for linear probing and for downstream tasks, however, a new kNN evaluation protocol is needed for multi-label classification tasks.
Dense prediction is a suitable evaluation metric for downstream application. However, the strength of SSL will be validated if only the decoder is trained for dense prediction and encoder is kept fixed. We envisage that labelled data particularly for multi-label and dense prediction tasks should mainly be used for linear, kNN and downstream evaluation, saving human effort and energy in data labelling. Our initial results (Figure~\ref{fig:vis}) demonstrated the potential for the possibility of building a representation for each of the concepts/objects in an image without any label.


{\small
\bibliographystyle{ieee_fullname}
\bibliography{egbib}
}

\appendix
\label{sec:append}

\clearpage
\section{Ablation Studies}
\label{sec:Ablation}

Due to limited resources, our ablation studies are conducted on 5\% of ImageNet-1K for training, and evaluated on the full validation set of ImageNet-1K. The models are pre-trained for $800$ epochs employing the ViT-S/16 architecture as the backbone of the student and the teacher of MC-SSL0.0.

\begin{figure}[h]
    \centering
    \includegraphics[width=\linewidth]{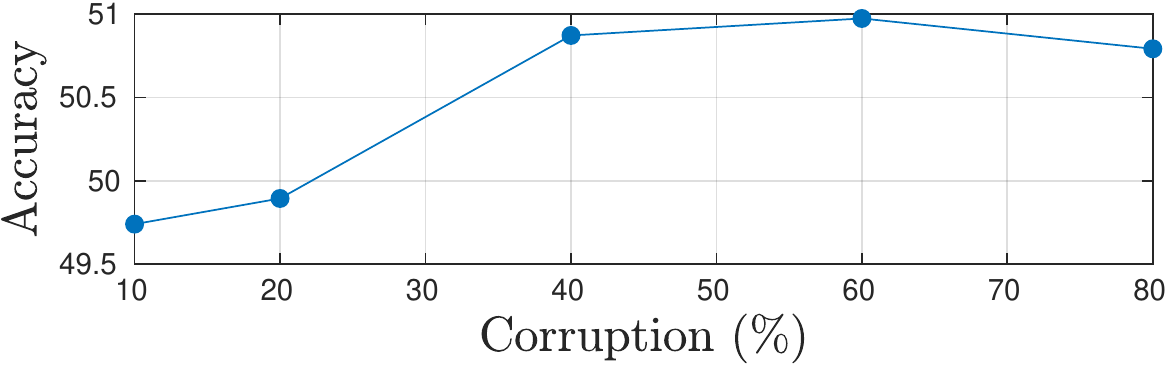}
    \caption{Effect of percentage of the corruption. }
    \label{fig:fine_corrupt}
\end{figure}

\noindent
\textbf{Effect of Image Corruption.}
Figure \ref{fig:fine_corrupt} shows the top-1 accuracy on fine-tuning when pre-trained with different corruption percentages, i.e. upto 10\%, 20\%, 40\%, 60\%, and 80\%. We found that the optimal ratio is between 40\% to 80\%. This behaviour was expected as the masking encourages the network to learn semantic information from the uncorrupted patches surrounding the masked tokens in order to recover the missing information. 

\begin{figure}[h]
    \centering
    \includegraphics[width=\linewidth]{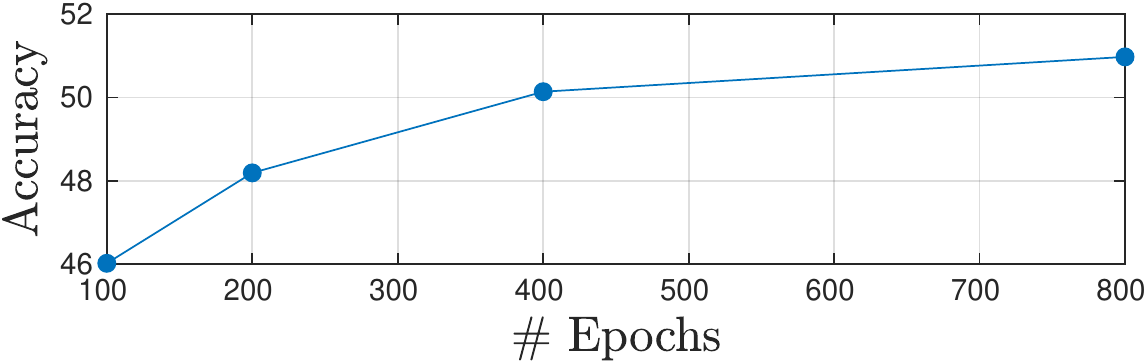}
    \caption{Effect of longer training. }
    \label{fig:long_tr}
\end{figure}

\noindent
\textbf{Effect of Longer Training.} Figure \ref{fig:long_tr} shows the top-1 accuracy when the model is pre-trained for 100, 200, 400, and 800 epochs. We found that longer pre-training improves the performance of MC-SSL0.0, where the accuracy is steadily improving even after 800 epochs of pre-training.

\begin{figure*}[h!]
\centering
\begin{subfigure}[t]{\textwidth}
\centering
\includegraphics[width=0.45\textwidth]{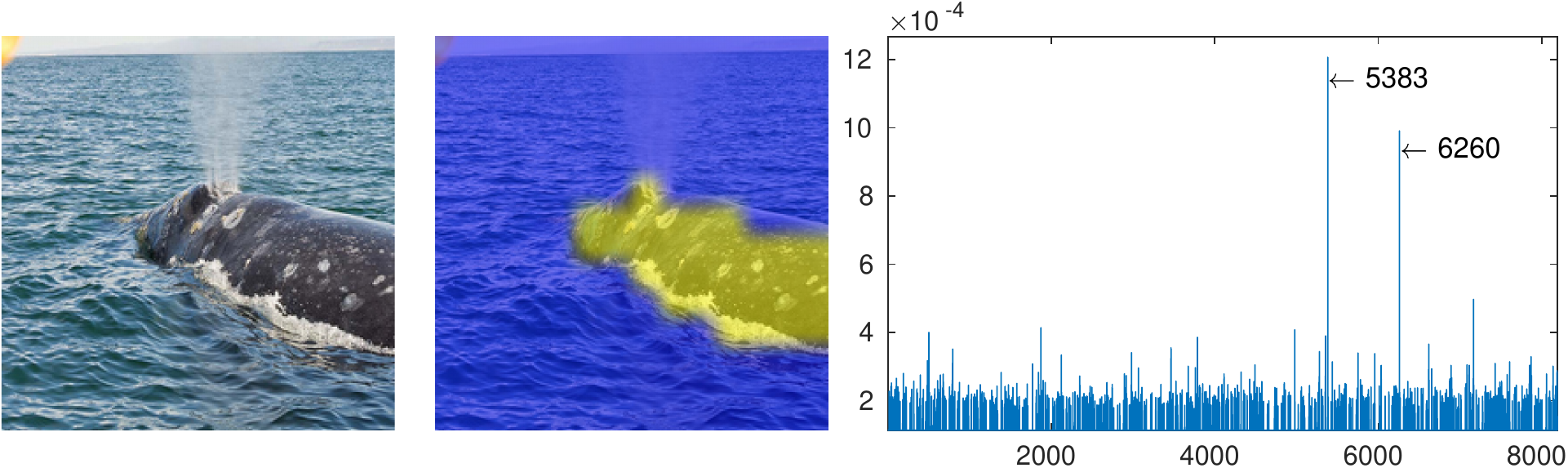}
\hspace{0.9cm}
\includegraphics[width=0.45\textwidth]{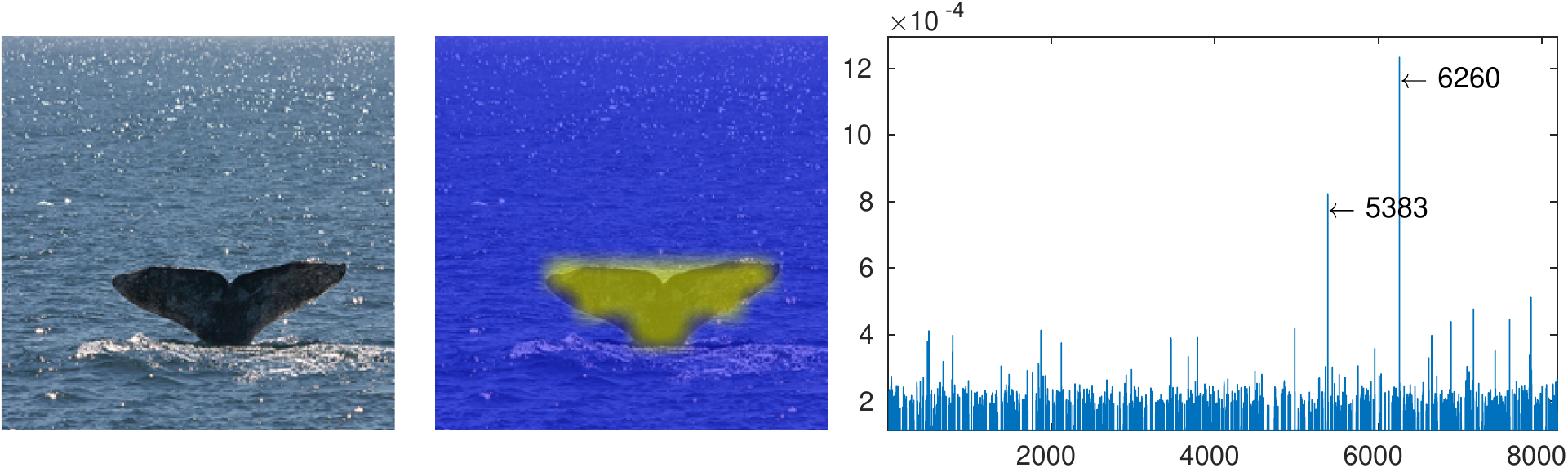}
\end{subfigure}
\begin{subfigure}[t]{\textwidth}
\centering
\includegraphics[width=0.45\textwidth]{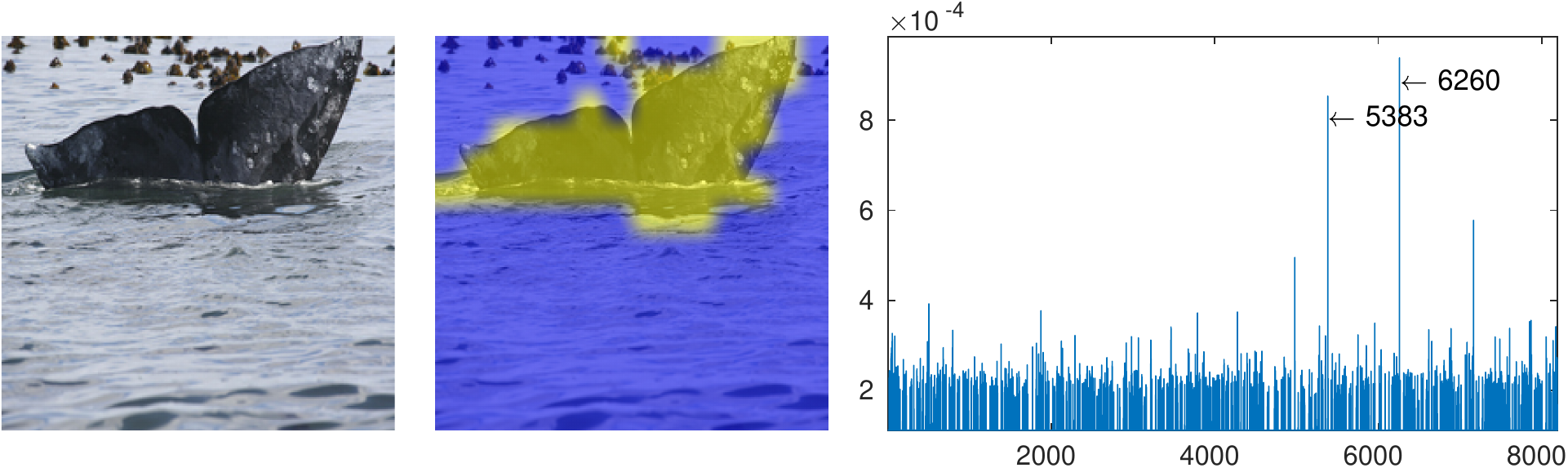}
\hspace{0.9cm}
\includegraphics[width=0.45\textwidth]{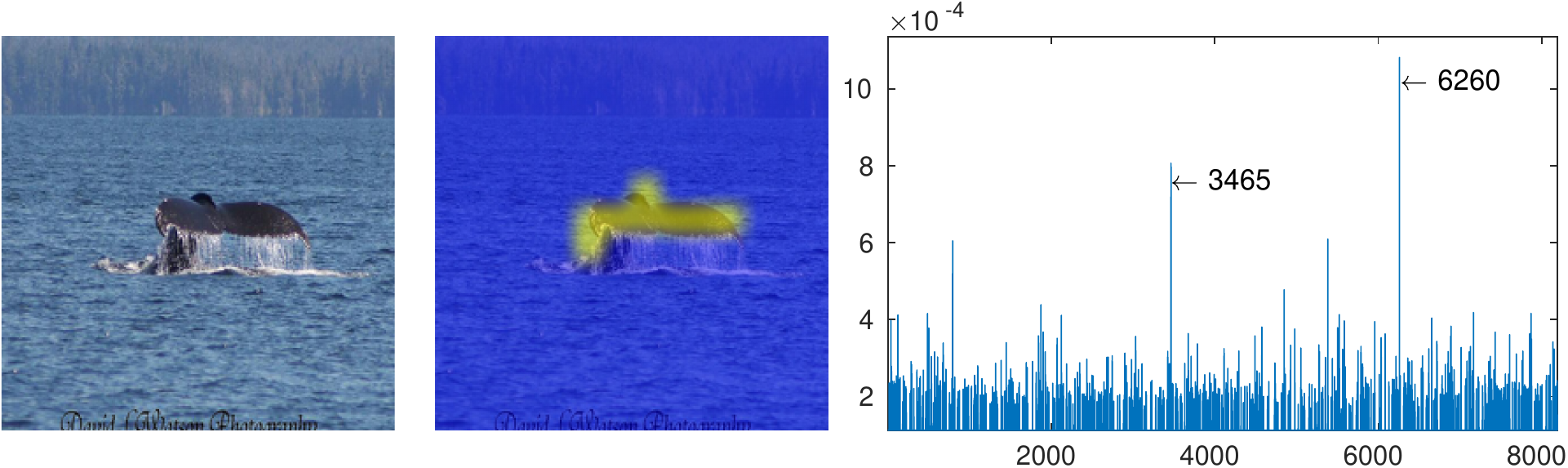}
\end{subfigure}
\begin{subfigure}[t]{\textwidth}
\centering
\includegraphics[width=0.45\textwidth]{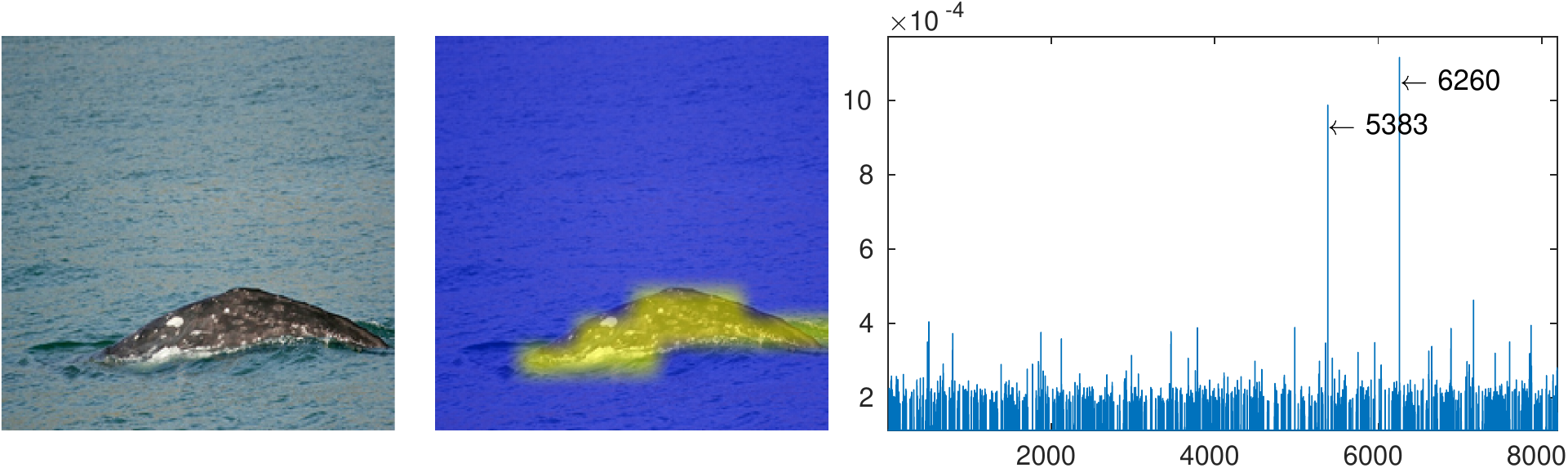}
\hspace{0.9cm}
\includegraphics[width=0.45\textwidth]{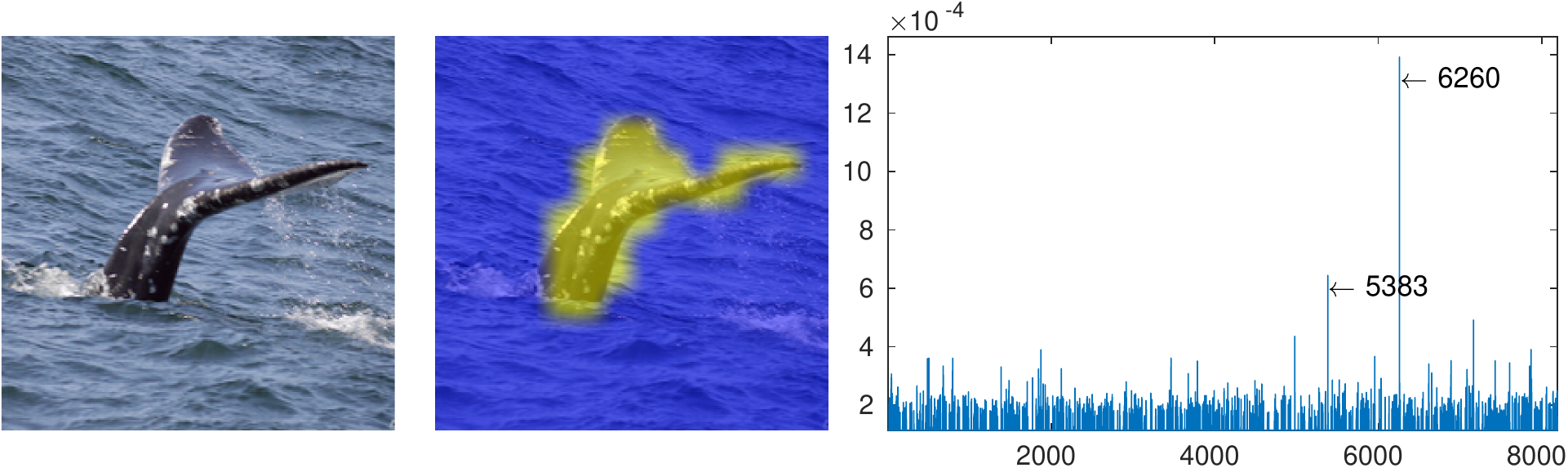}
\end{subfigure}
\begin{subfigure}[t]{\textwidth}
\centering
\includegraphics[width=0.45\textwidth]{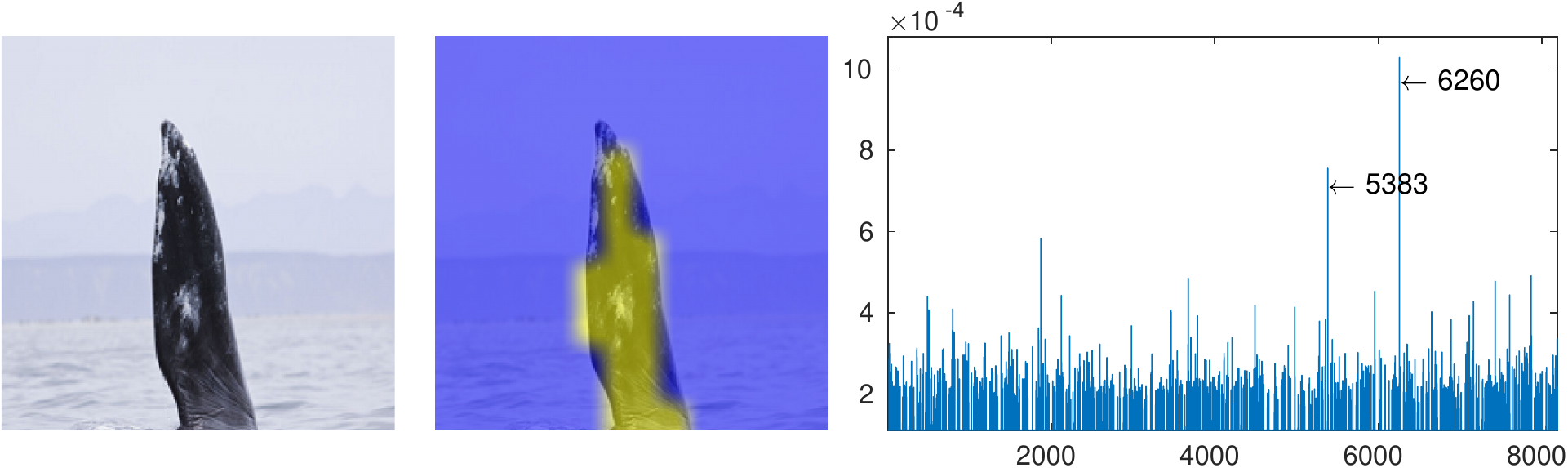}
\hspace{0.9cm}
\includegraphics[width=0.45\textwidth]{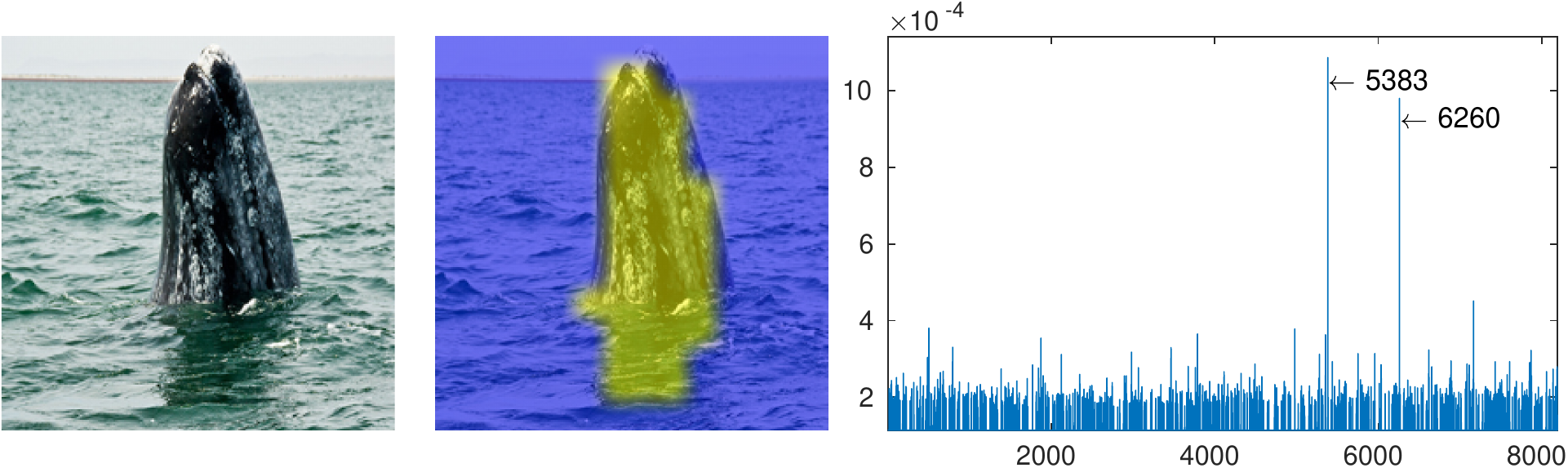}
\end{subfigure}
\begin{subfigure}[t]{\textwidth}
\centering
\includegraphics[width=0.3\textwidth]{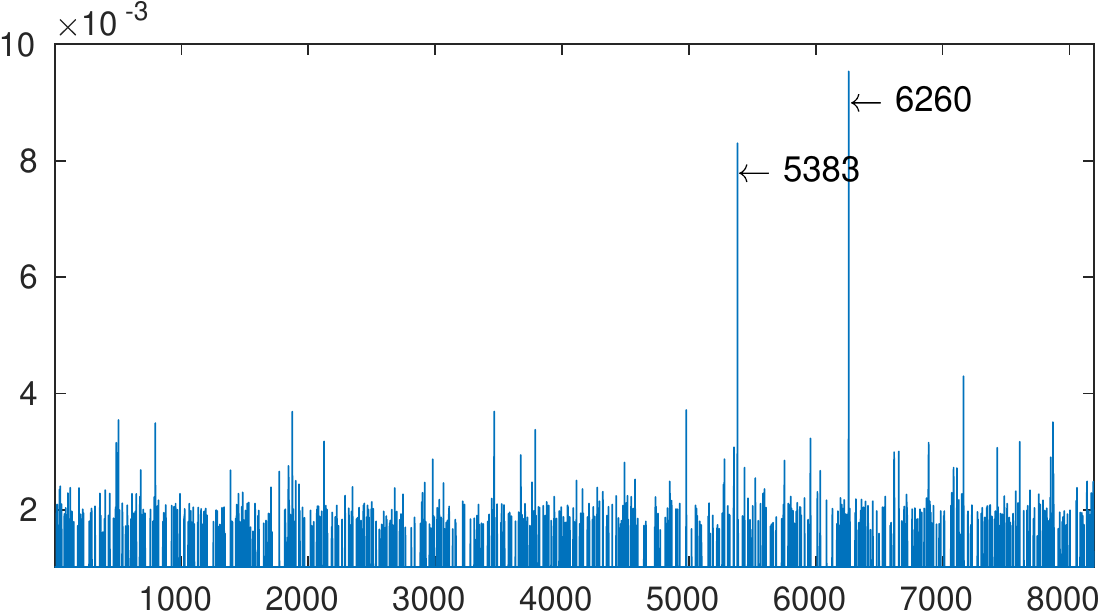}
\caption{Average of the probabilities corresponding to the dominant concept (yellow patches)}
\end{subfigure}
\caption{The image is clustered into two concepts which are self learnt. First self-learnt cluster is attending the dominant object (shown by yellow colour in centre column) and other is focusing on the environment (shown by blue colour in centre column). 
The bar plots are showing the probability of the learnt patch concepts corresponding to the dominant object. As can be seen the learnt patch concepts are consist across different instances of semantic concept in different images enforcing the thesis of MC-SSL0.0.}
\label{fig:whaleVisual}
\end{figure*}

\begin{figure*}[h!]
\centering
\begin{subfigure}[t]{\textwidth}
\centering
\includegraphics[width=0.45\textwidth]{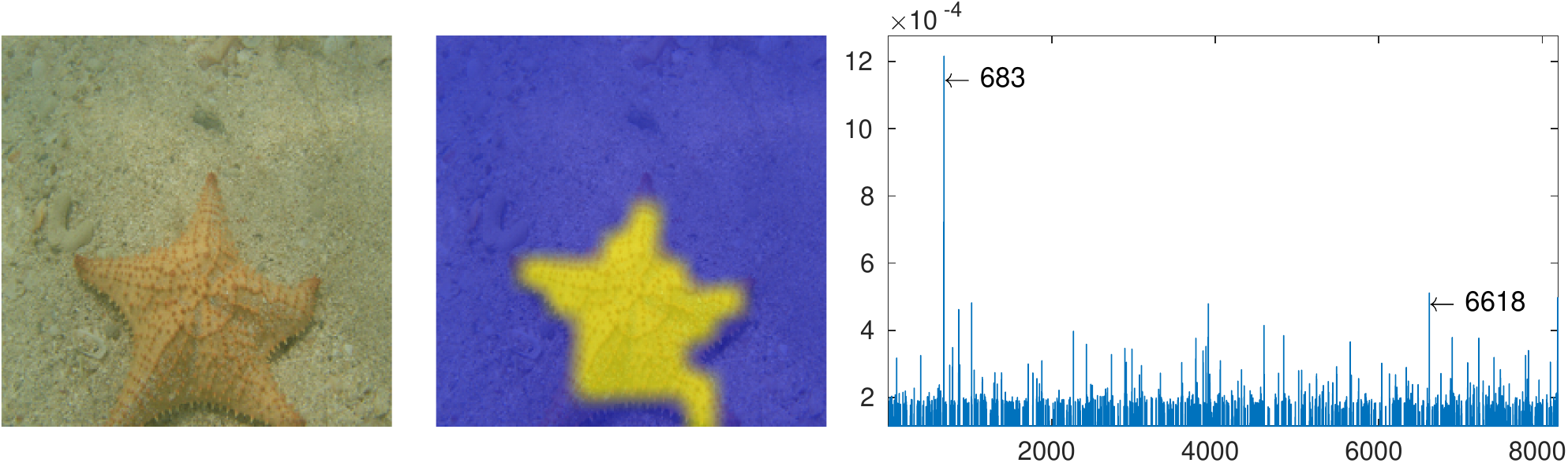}
\hspace{0.9cm}
\includegraphics[width=0.45\textwidth]{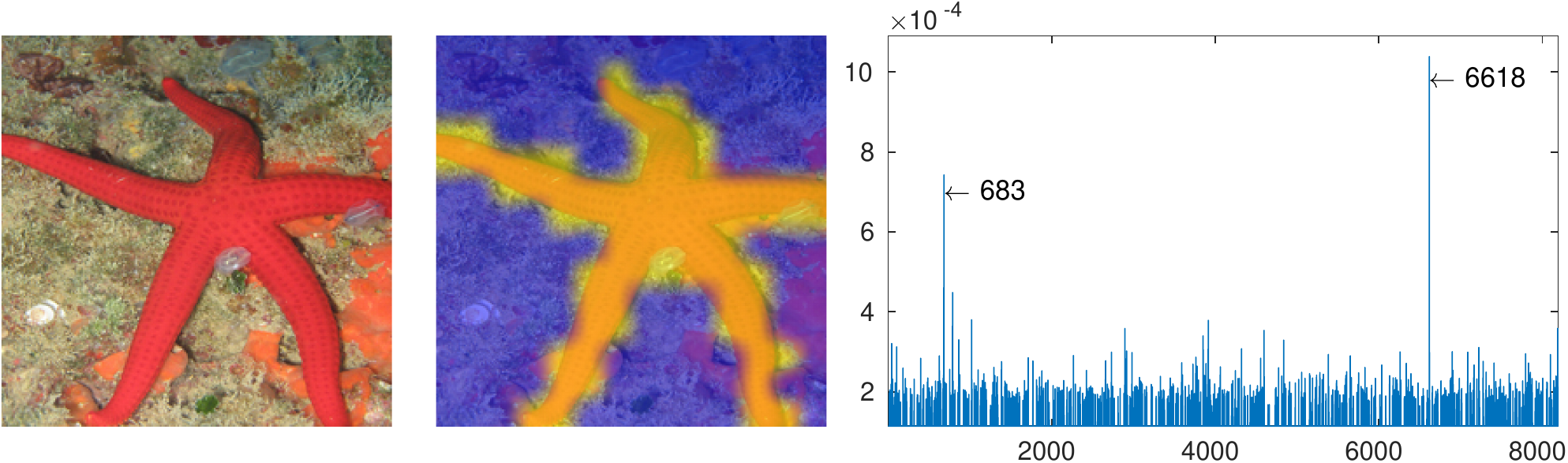}
\end{subfigure}
\begin{subfigure}[t]{\textwidth}
\centering
\includegraphics[width=0.45\textwidth]{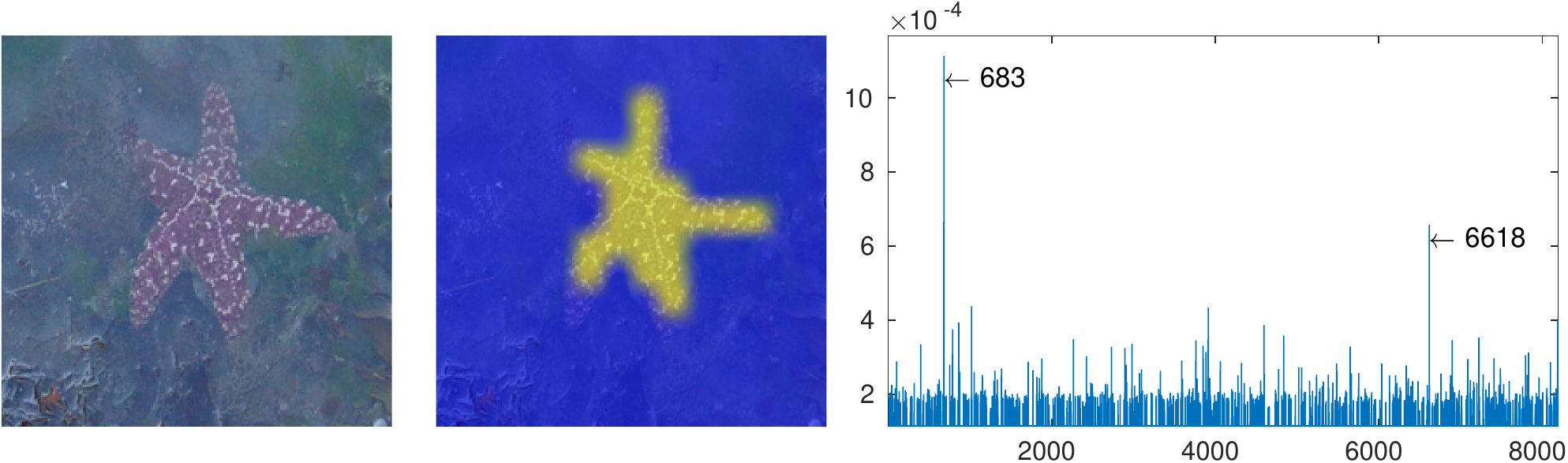}
\hspace{0.9cm}
\includegraphics[width=0.45\textwidth]{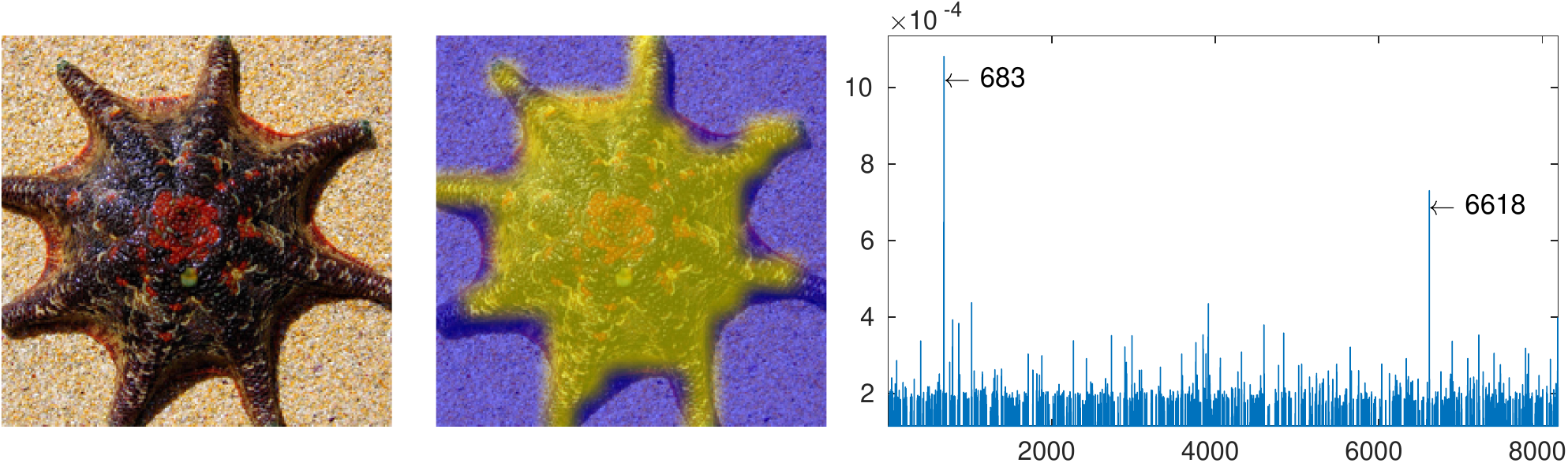}
\end{subfigure}
\begin{subfigure}[t]{\textwidth}
\centering
\includegraphics[width=0.45\textwidth]{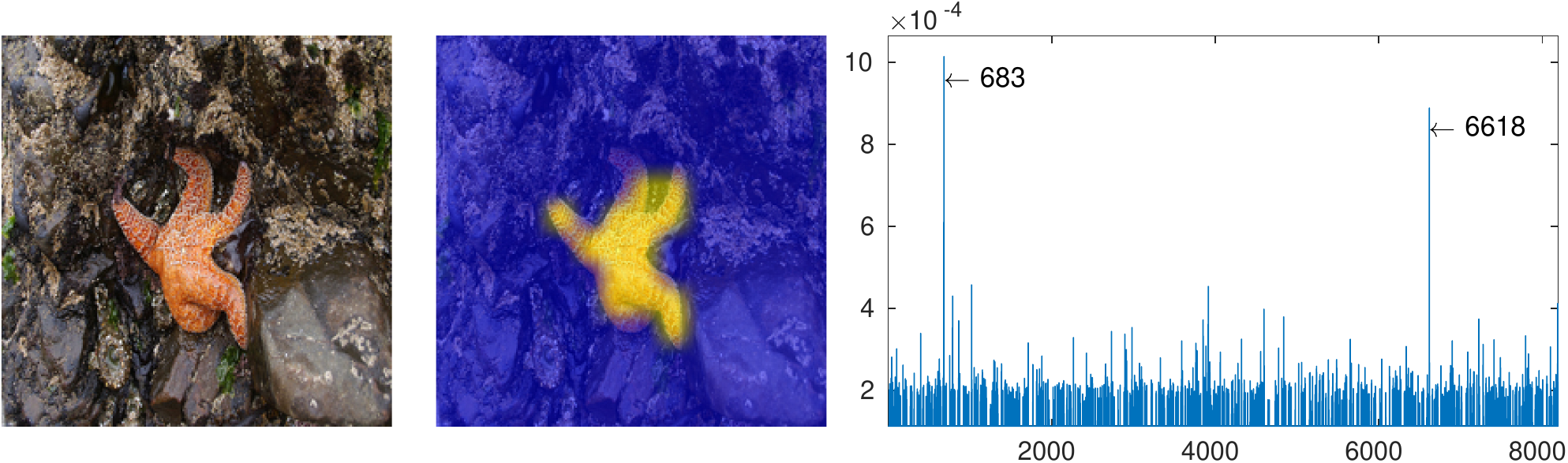}
\hspace{0.9cm}
\includegraphics[width=0.45\textwidth]{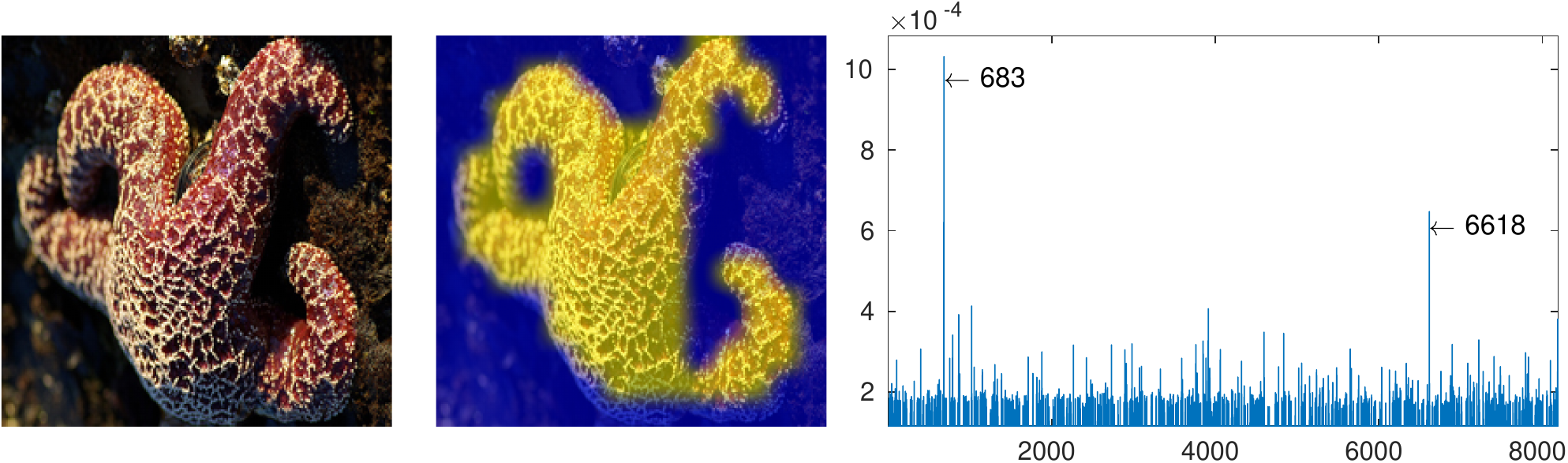}
\end{subfigure}
\begin{subfigure}[t]{\textwidth}
\centering
\includegraphics[width=0.45\textwidth]{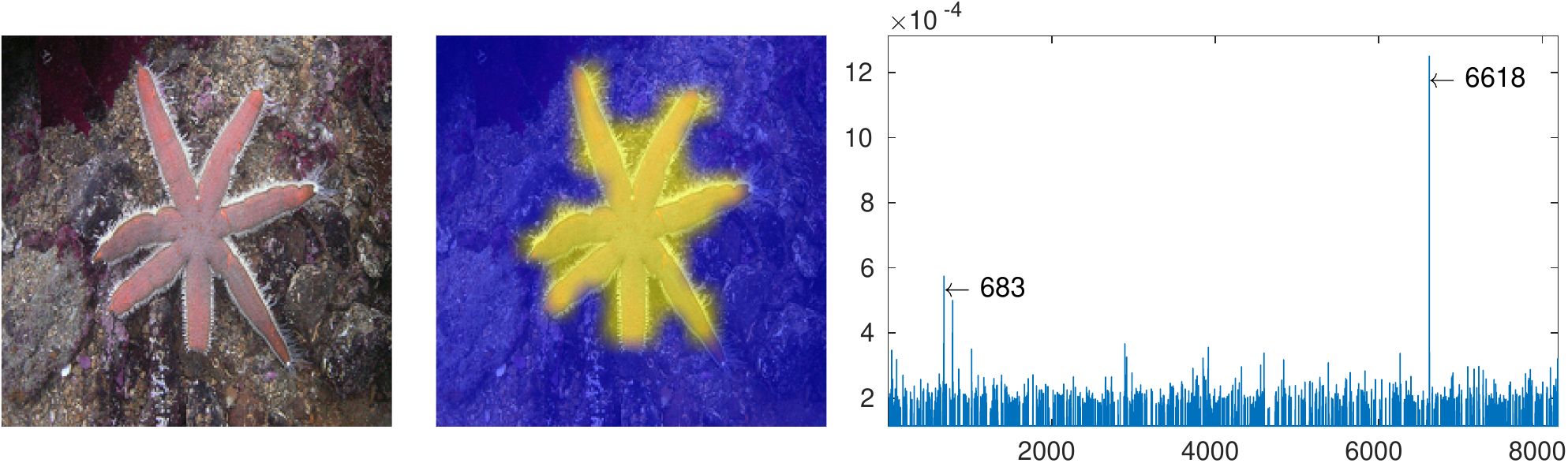}
\hspace{0.9cm}
\includegraphics[width=0.45\textwidth]{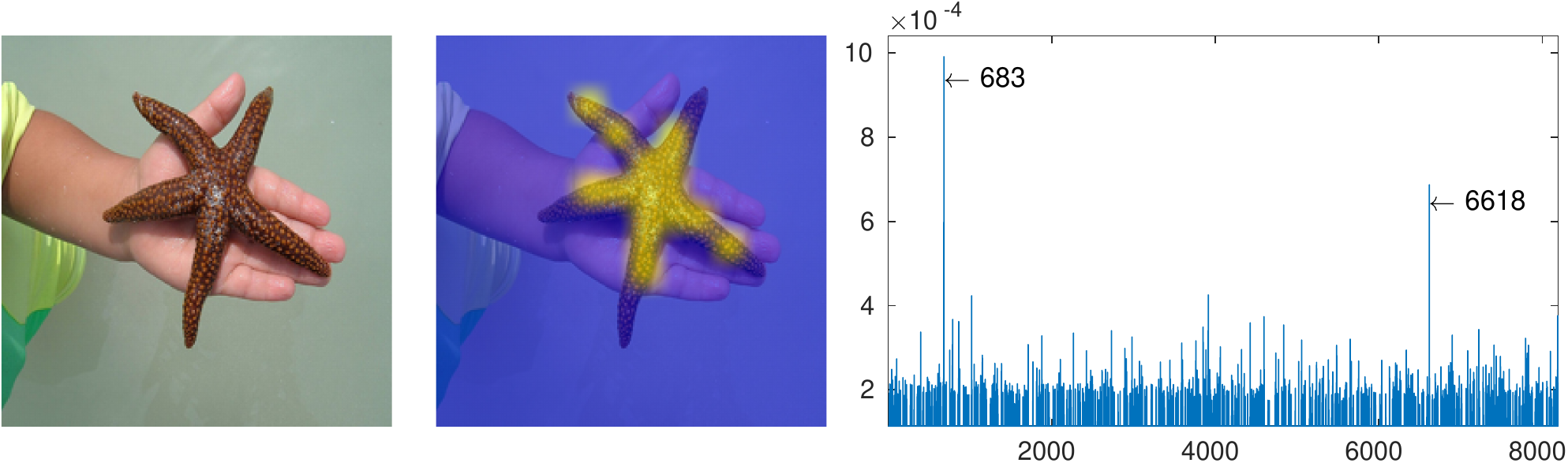}
\end{subfigure}
\begin{subfigure}[t]{\textwidth}
\centering
\includegraphics[width=0.3\textwidth]{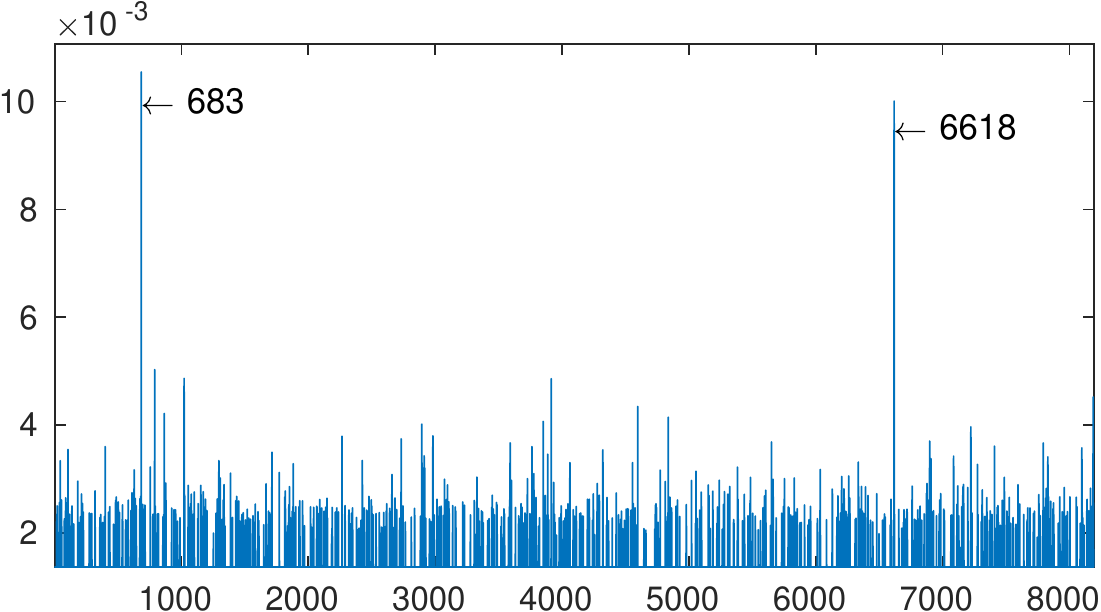}
\end{subfigure}
\caption{The image is clustered into two concepts which are self learnt. First self-learnt cluster is attending the dominant object (shown by yellow colour in centre column) and other is focusing on the environment (shown by blue colour in centre column). 
The bar plots are showing the probability of the learnt patch concepts corresponding to the dominant object. As can be seen the learnt patch concepts are consist across different instances of semantic concept in different images enforcing the thesis of MC-SSL0.0.}
\label{lastfig}
\end{figure*}

\noindent
\textbf{Effect of Different Components of MC-SSL0.0.}
In this set of experiments, we used 10\% of ImageNet dataset for pre-training and validate the effect of each component of MC-SSL0.0 by finetuning the pre-trained model on several downstream tasks.
In Table \ref{tbl:ablation}, we show the effect of using only the patch reconstruction objective (MC-SSL0.0 [PR]), using only the patch classification objective (MC-SSL0.0 [PC]), and using both, patch classification and patch reconstruction, objectives (MC-SSL0.0 [PC + PR]). Note that the teacher network is only required for the patch classification task, therefore, it is not included in the case of MC-SSL0.0 [PR].
Group Masked Model Learning (GMML) based PR 
learn the knowledge about an object/concept (that is properties such as colour, texture and structure, as well as context) in order to reconstruct, as well as to recover the distorted data-tokens by using available information in unmasked data tokens on the object and its surroundings. This encourages all the data-tokens on an object to have similar representation to each other and incorporate local context in transformers as shown in Figure~\ref{fig:vis}. 
The PR can server as a strong self-supervised pretext task for vision modality. 
It has the advantage of simplicity and does not need teacher student framework using momentum based update of teacher. Also PR does not require careful selection of parameters such as temperature, centring and other careful engineering designs. 
The ultimate role of the auxiliary but complementary task of learning a patch classifier is to assign a pseudo-semantic label to a group of context aware data tokens covering an object. 
We note that patch concept learning on its own can also be used as a strong self-supervised pretext task. However, in order for it to performs well it need careful engineering design efforts, like centring, momentum based encoder for teacher, careful selection of temperature. 
Nevertheless, Our conjecture is that learning local semantics using GMML base PR and learning pseudo labels for patches encourages data tokens on similar objects within an image and between images to belong to the same pseudo class promoting intra and inter image concept consistency as shown in Figure~\ref{fig:whaleVisual} to Figure~\ref{lastfig}.

We found that the performance is improved by combining patch reconstruction and patch classification objectives in the proposed MC-SSL0.0, especially in the case of Cars dataset where the accuracy jumped from 83.93\% to 86.15\%.

In summary, both the patch recontruction and patch classification losses with GMML on their own provide as a means of self-supervision and can provided an effective starting point for efficient downstream task finetuning. Further improvements are obtained by combiningn the two in MC-SSL0.0 framework. One of the main objectives of MC-SSL is to explore the possibility of learning representation for each object in an image which is consistant across the dataset. Initial results and visualisation show a promising horizon for building upon MC-SSL0.0 framework.  

\begin{table}[t]
\centering
\caption{Transfer learning by finetuning pretrained Self-supervised models on different downstream tasks. Self-supervised models are trained using ViT-S/16 model on $10\%$ of ImageNet dataset, followed by finetuning on downstream tasks.}
\label{tbl:ablation}
\resizebox{\linewidth}{!}{
\begin{tabular}{l|c|c|c|c}
\hline
       & CIFAR10~     & CIFAR100   & ~~Cars~~    & Flowers~  \\ \hline
\itshape{\textcolor{darkgray}{Random Init.}} & \itshape{\textcolor{darkgray}{91.42}}    
& \itshape{\textcolor{darkgray}{70.14}}& \itshape{\textcolor{darkgray}{10.67}}  
& \itshape{\textcolor{darkgray}{54.04}}\\ \hline
& \multicolumn{4}{c}{w/o multi-crop} \\ \cline{2-5} 
MC-SSL0.0 [PR]                           & 97.19 & 81.98 & 76.78  & 88.21\\ 
MC-SSL0.0 [PC]                           & 97.77 & 84.25 & 83.93  & 94.89 \\ 
MC-SSL0.0 [PC + PR] & \textbf{97.82}&\textbf{84.98}&\textbf{86.15}&\textbf{95.56}\\\hline
\end{tabular}
}
\end{table}

\section{Visualisation}
\label{sec:visualisation}

\begin{figure*}[h!]
\centering
\begin{subfigure}[t]{\textwidth}
\centering
\includegraphics[width=0.45\textwidth]{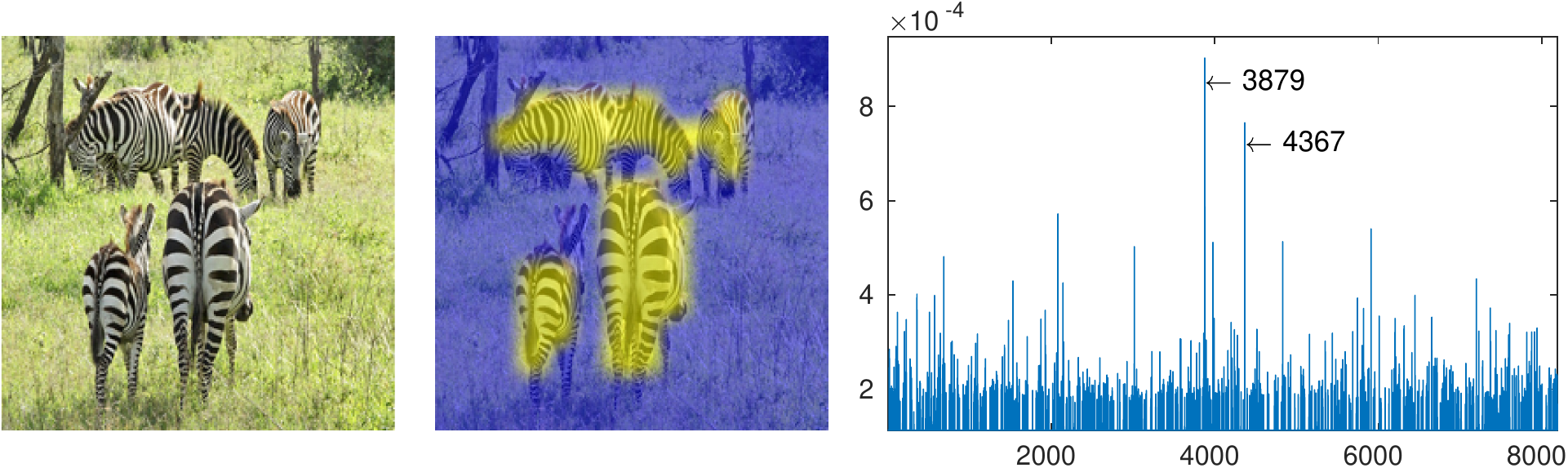}
\hspace{0.9cm}
\includegraphics[width=0.45\textwidth]{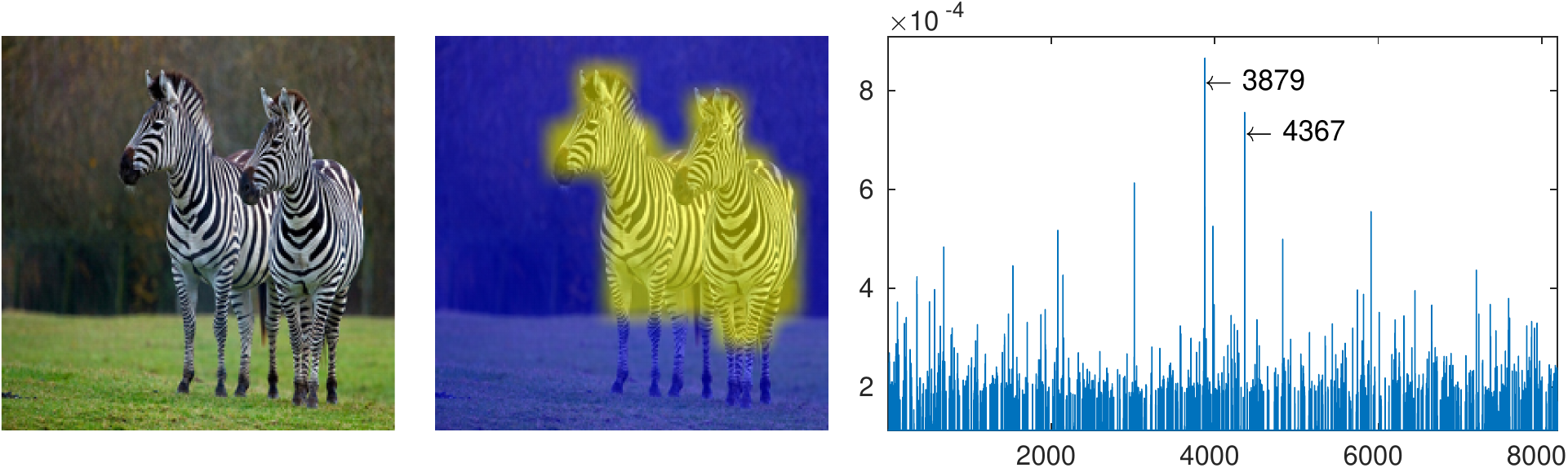}
\end{subfigure}
\begin{subfigure}[t]{\textwidth}
\centering
\includegraphics[width=0.45\textwidth]{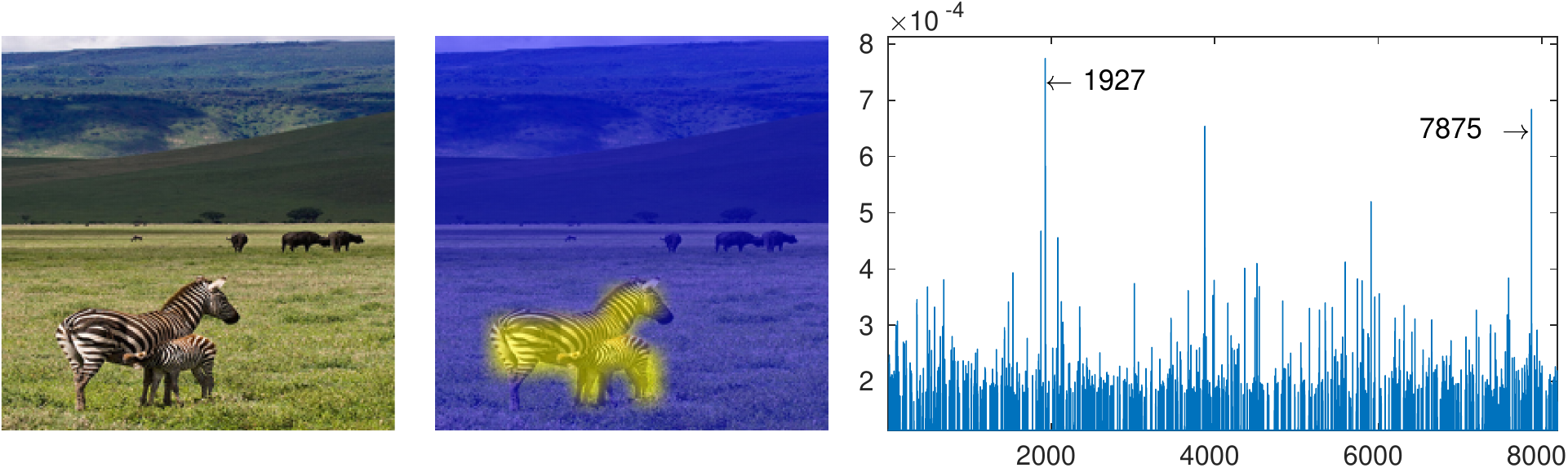}
\hspace{0.9cm}
\includegraphics[width=0.45\textwidth]{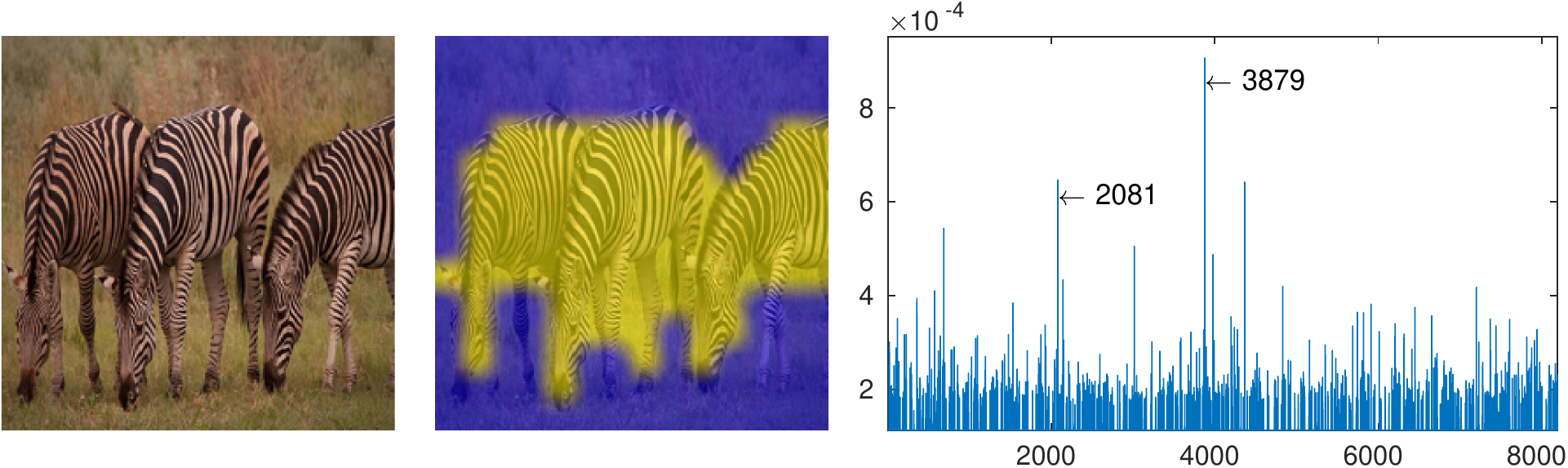}
\end{subfigure}
\begin{subfigure}[t]{\textwidth}
\centering
\includegraphics[width=0.45\textwidth]{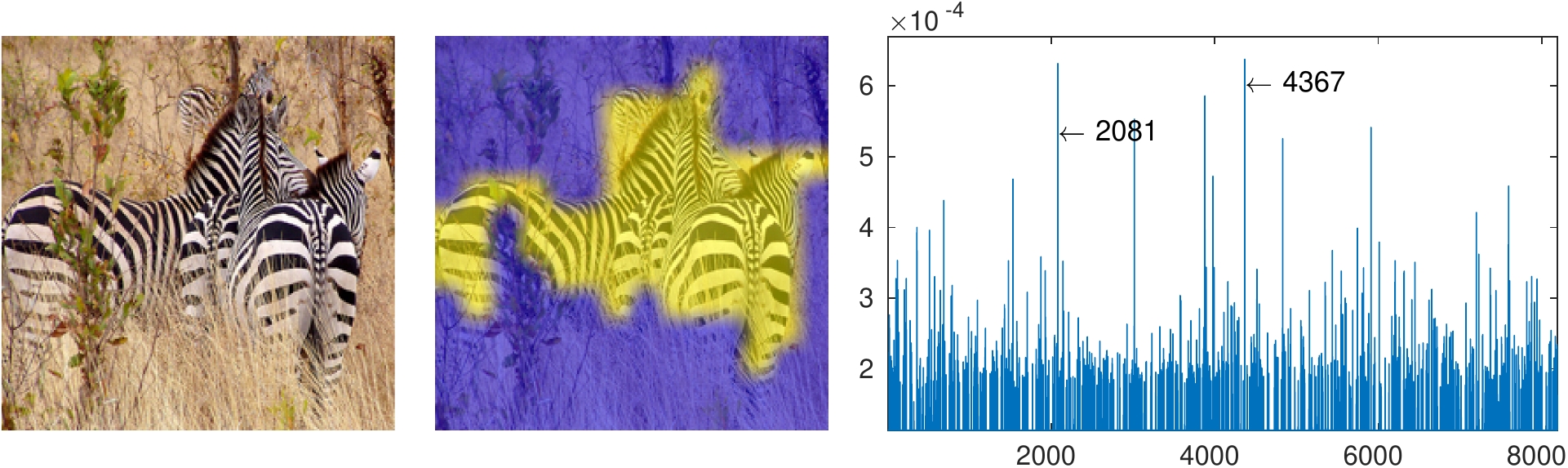}
\hspace{0.9cm}
\includegraphics[width=0.45\textwidth]{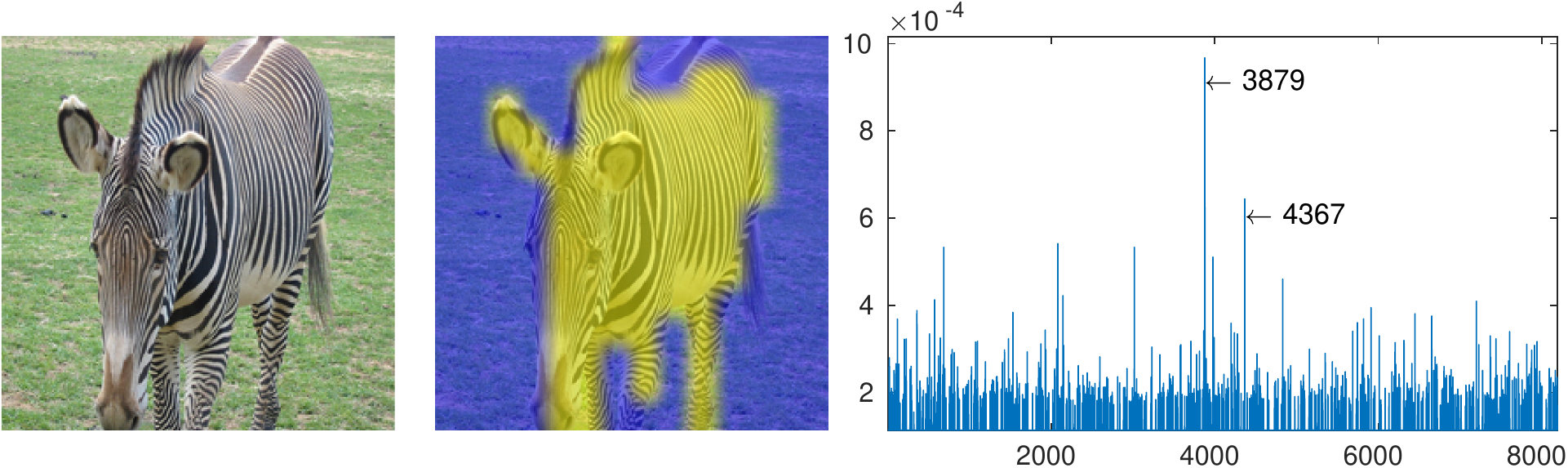}
\end{subfigure}
\begin{subfigure}[t]{\textwidth}
\centering
\includegraphics[width=0.45\textwidth]{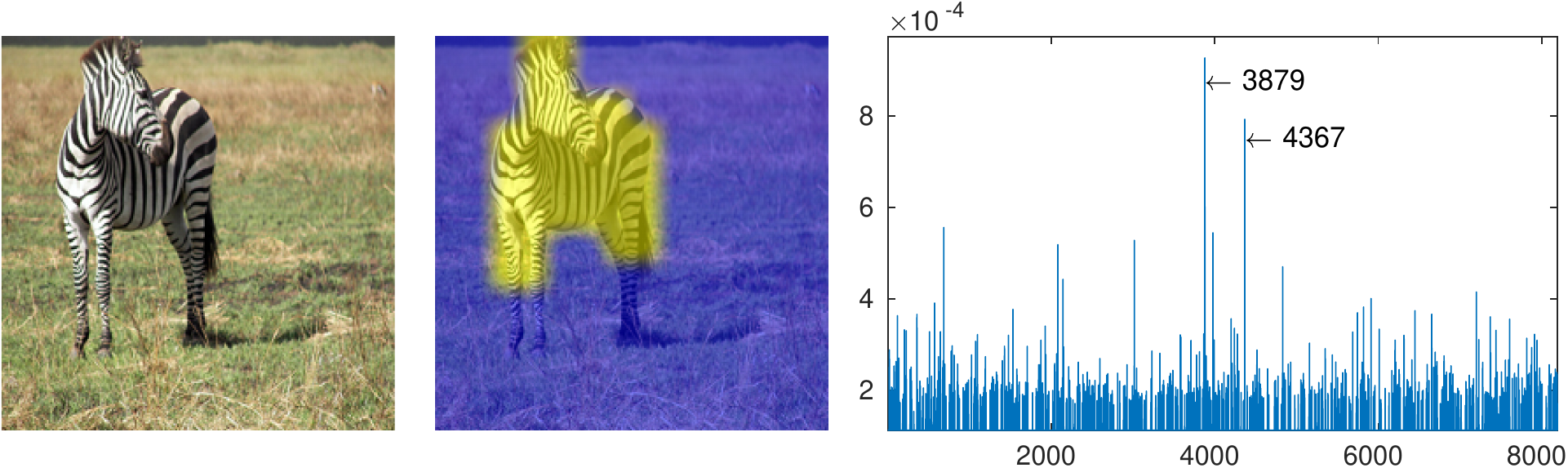}
\hspace{0.9cm}
\includegraphics[width=0.45\textwidth]{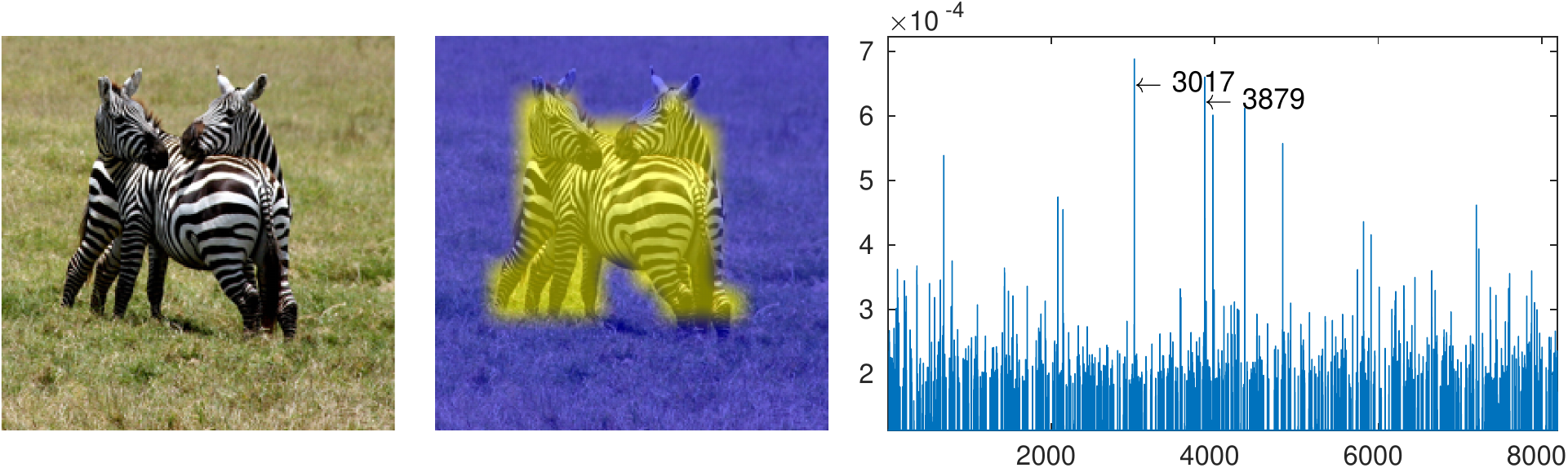}
\end{subfigure}
\begin{subfigure}[t]{\textwidth}
\centering
\includegraphics[width=0.3\textwidth]{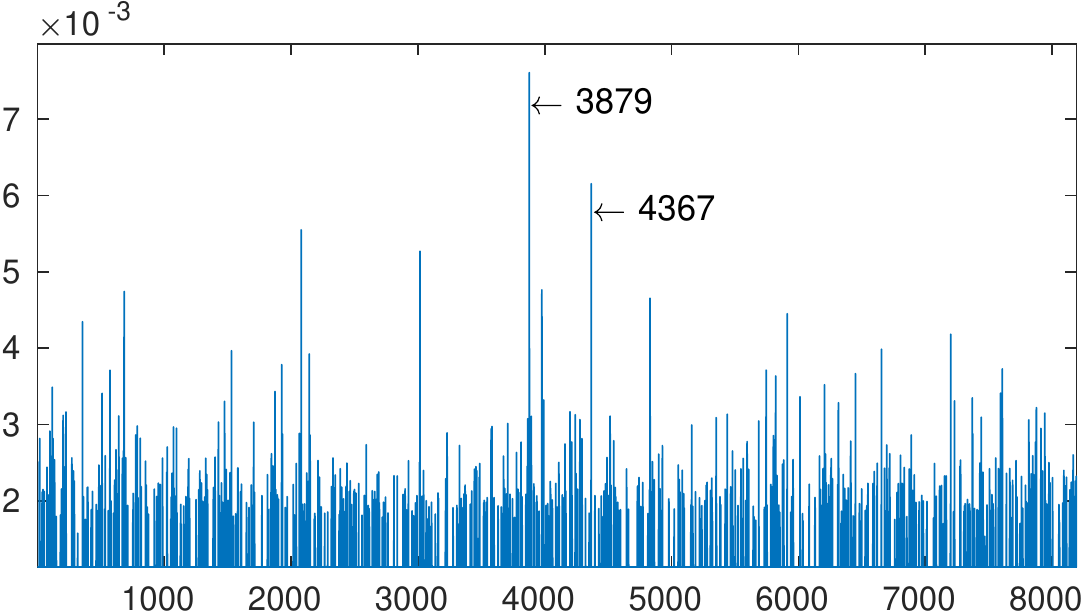}
\end{subfigure}
\caption{The image is clustered into two concepts which are self learnt. First self-learnt cluster is attending the dominant object (shown by yellow colour in centre column) and other is focusing on the environment (shown by blue colour in centre column). 
The bar plots are showing the probability of the learnt patch concepts corresponding to the dominant object. As can be seen the learnt patch concepts are consist across different instances of semantic concept in different images enforcing the thesis of MC-SSL0.0.}
\end{figure*}

\begin{figure*}[h!]
\centering
\begin{subfigure}[t]{\textwidth}
\centering
\includegraphics[width=0.45\textwidth]{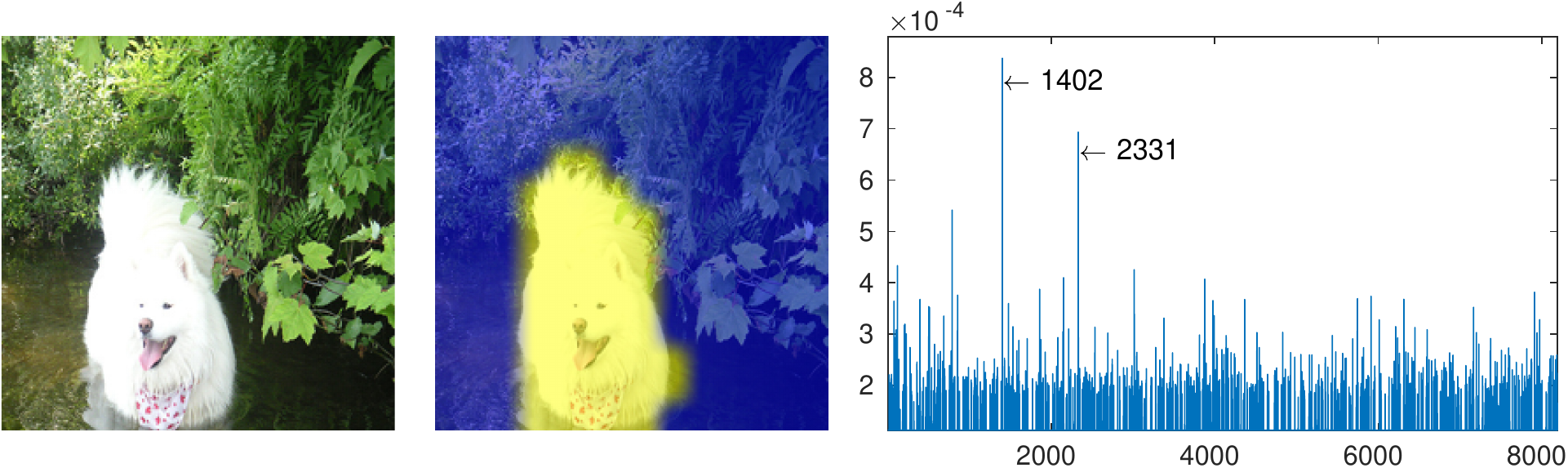}
\hspace{0.9cm}
\includegraphics[width=0.45\textwidth]{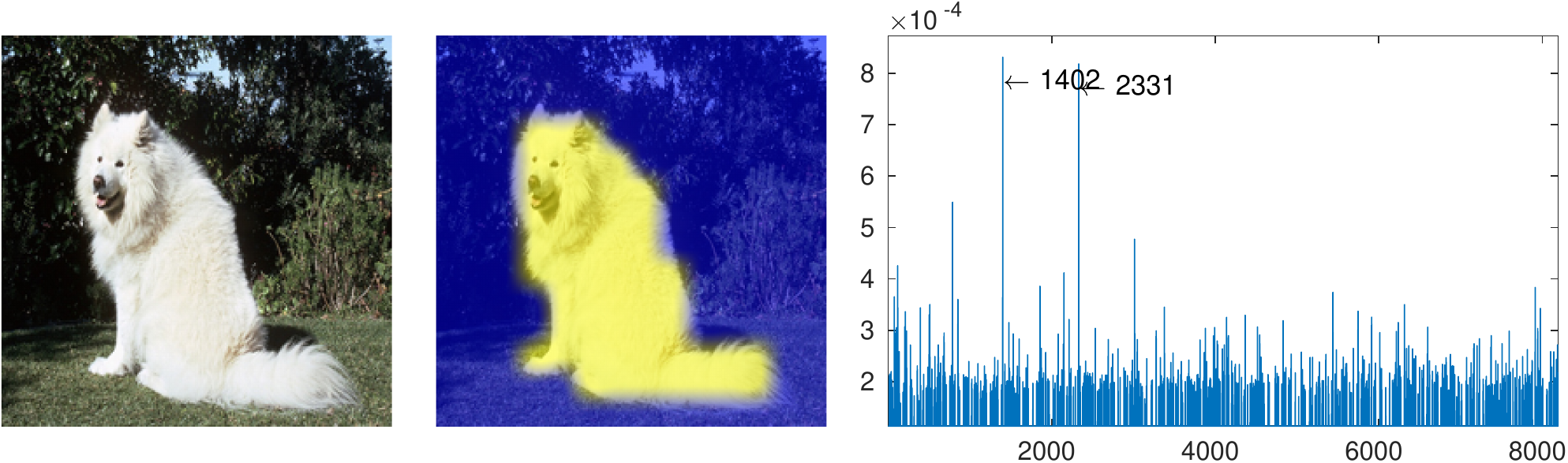}
\end{subfigure}
\begin{subfigure}[t]{\textwidth}
\centering
\includegraphics[width=0.45\textwidth]{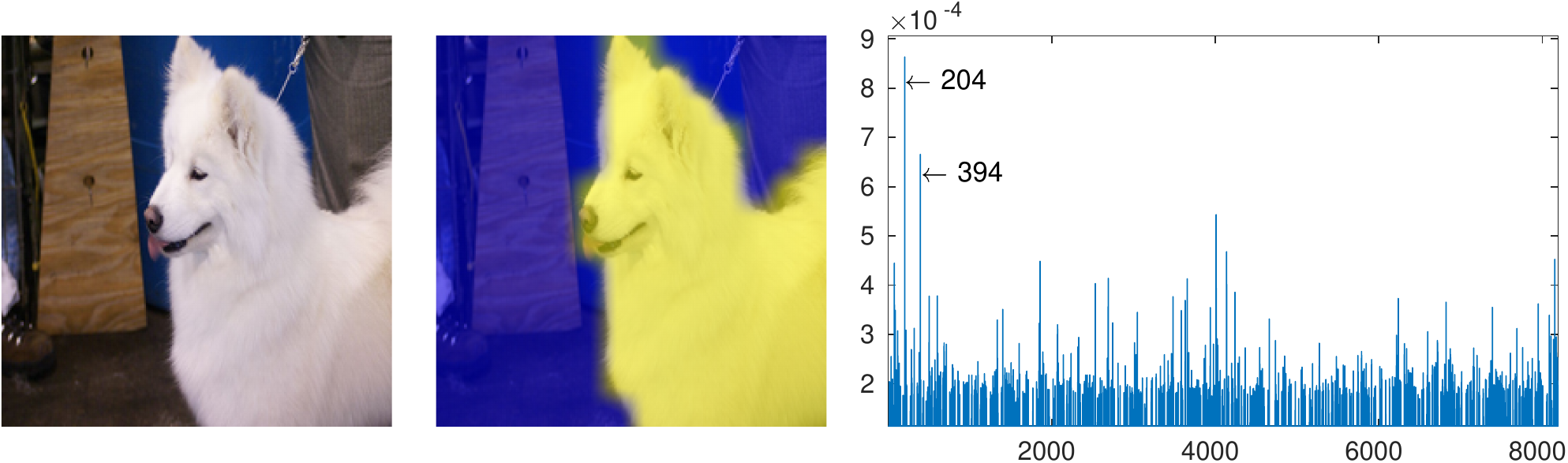}
\hspace{0.9cm}
\includegraphics[width=0.45\textwidth]{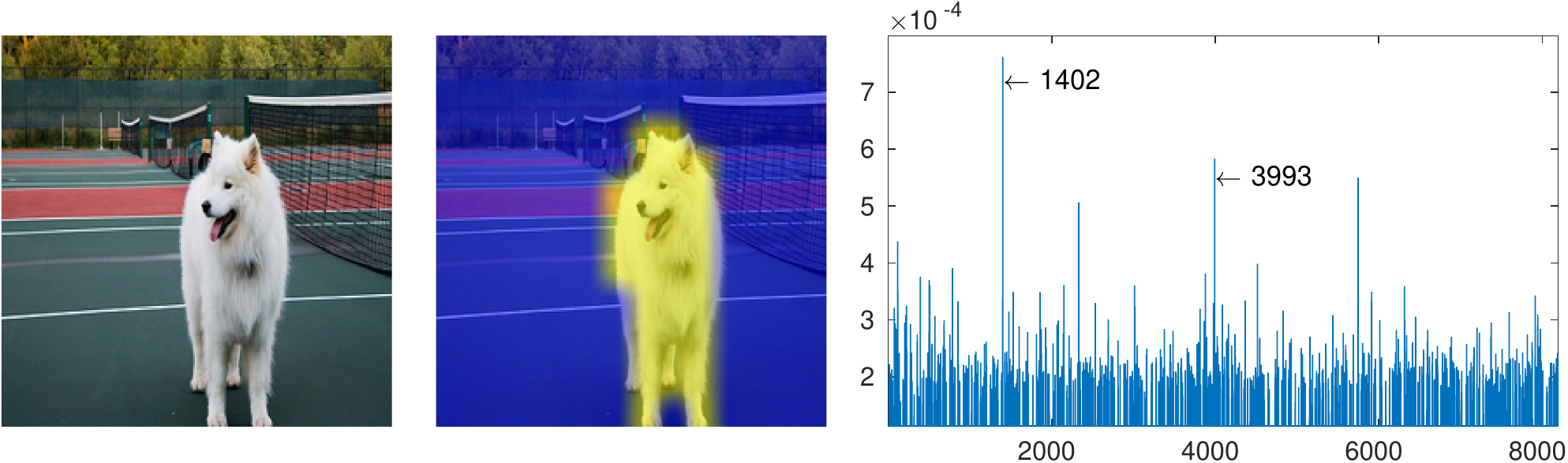}
\end{subfigure}
\begin{subfigure}[t]{\textwidth}
\centering
\includegraphics[width=0.45\textwidth]{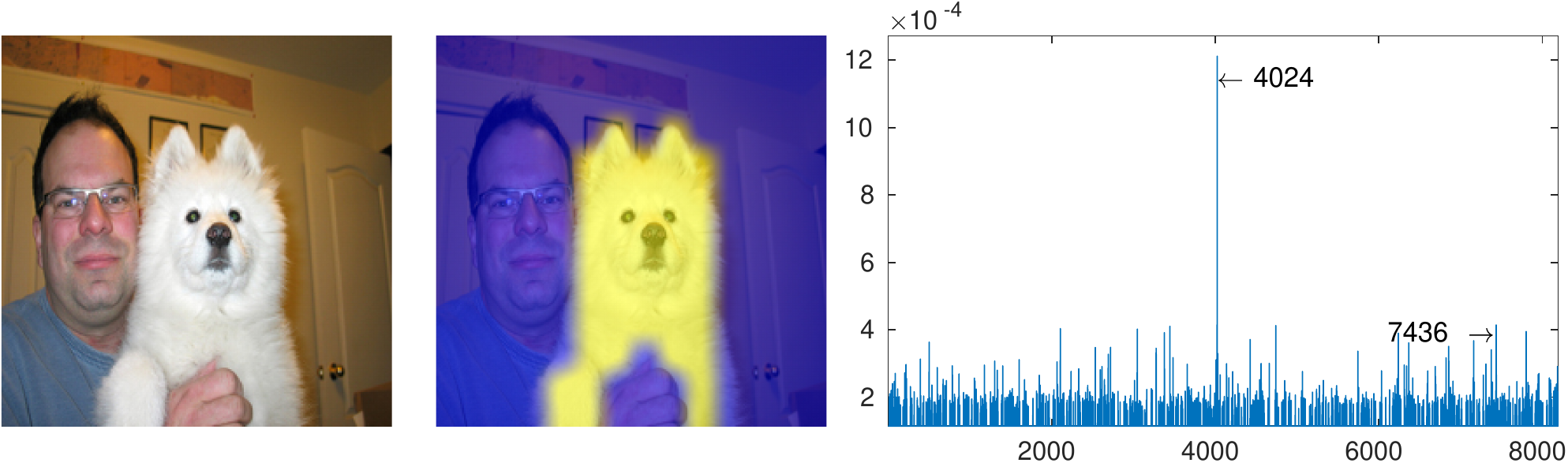}
\hspace{0.9cm}
\includegraphics[width=0.45\textwidth]{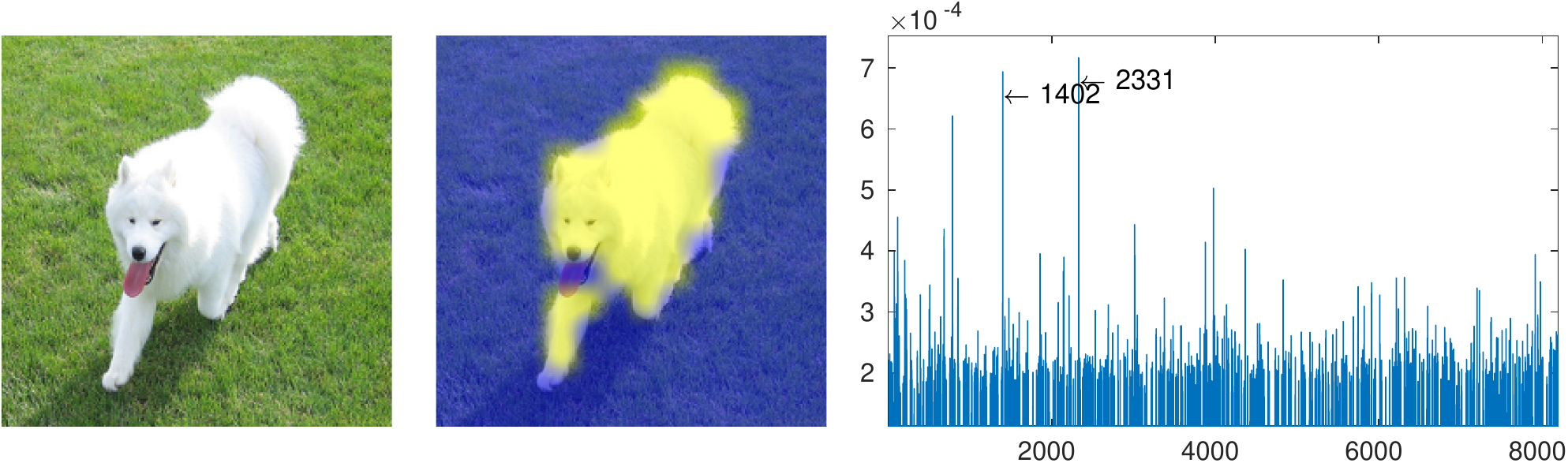}
\end{subfigure}
\begin{subfigure}[t]{\textwidth}
\centering
\includegraphics[width=0.45\textwidth]{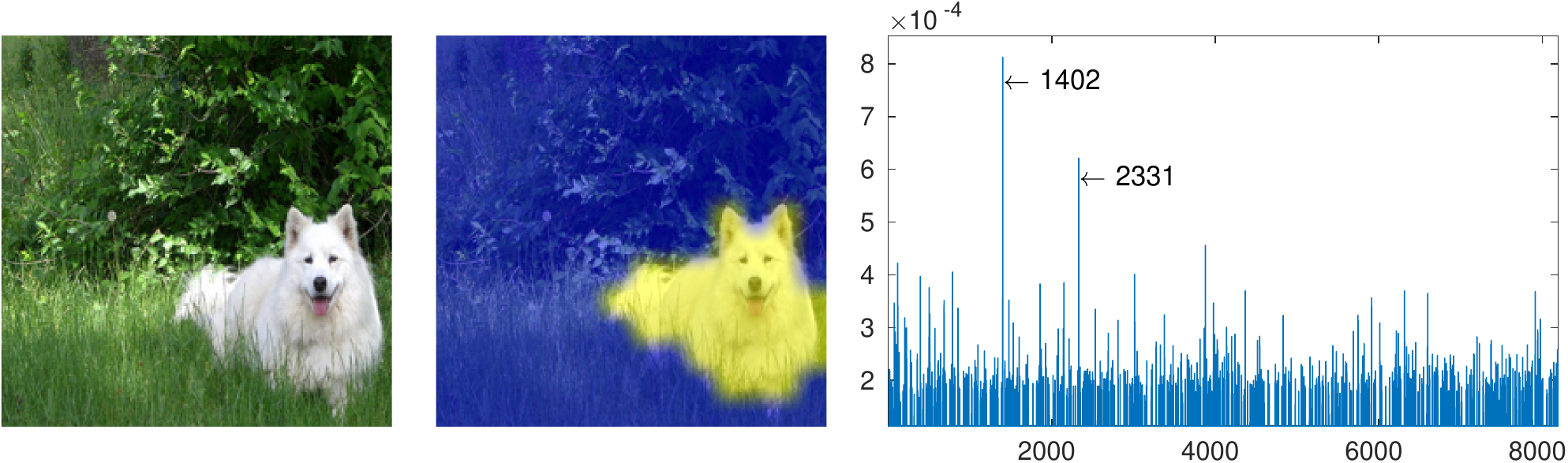}
\hspace{0.9cm}
\includegraphics[width=0.45\textwidth]{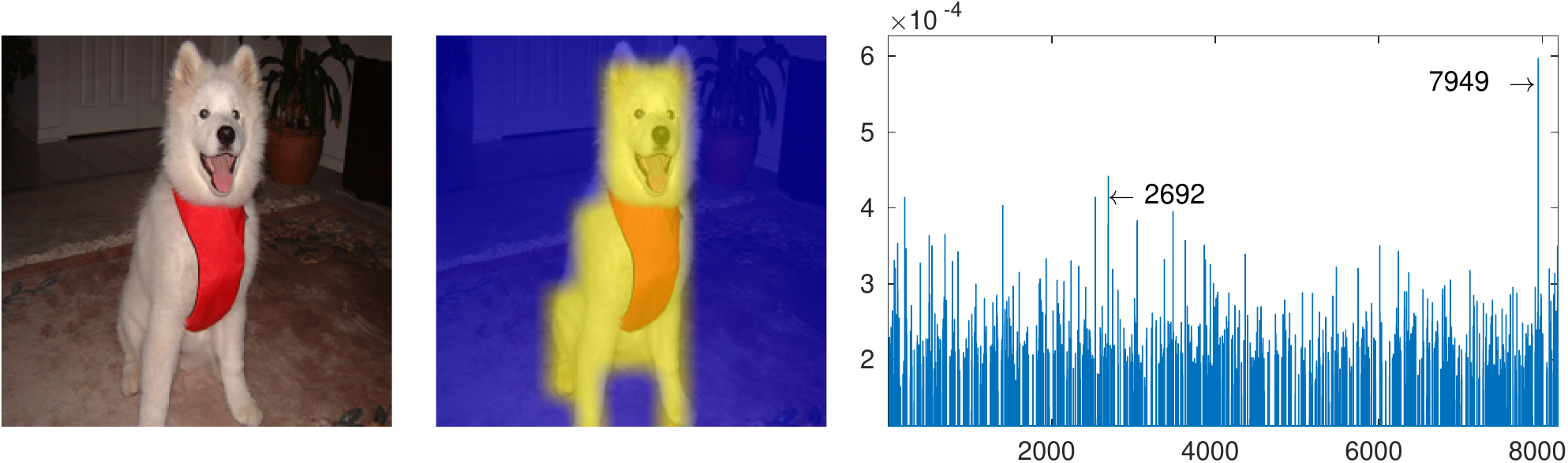}
\end{subfigure}
\begin{subfigure}[t]{\textwidth}
\centering
\includegraphics[width=0.3\textwidth]{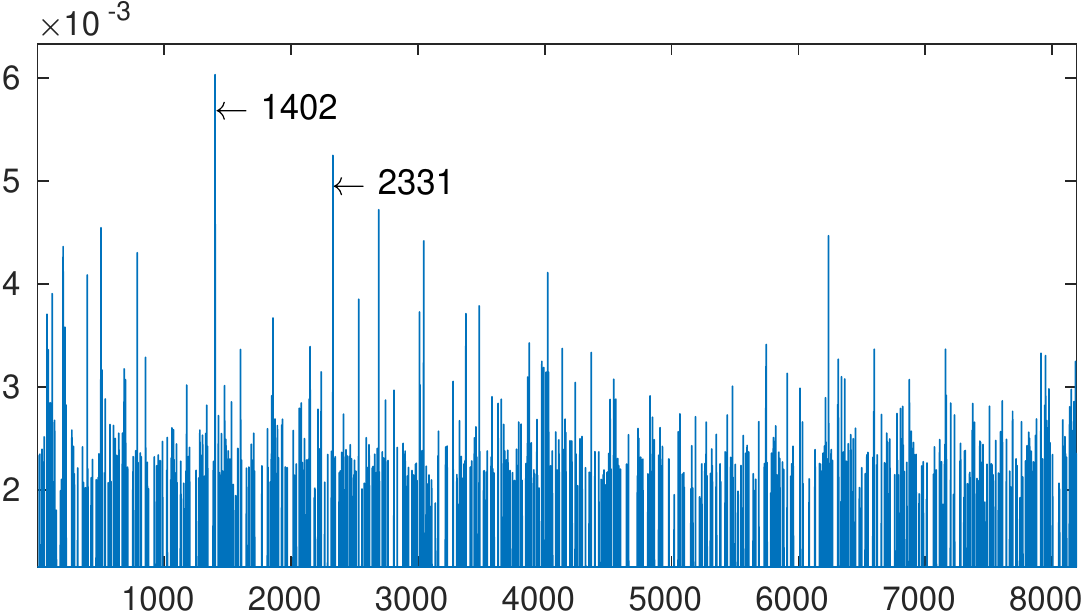}
\end{subfigure}
\caption{The image is clustered into two concepts which are self learnt. First self-learnt cluster is attending the dominant object (shown by yellow colour in centre column) and other is focusing on the environment (shown by blue colour in centre column). 
The bar plots are showing the probability of the learnt patch concepts corresponding to the dominant object. As can be seen the learnt patch concepts are consist across different instances of semantic concept in different images enforcing the thesis of MC-SSL0.0.}
\end{figure*}

\begin{figure*}[h!]
\centering
\begin{subfigure}[t]{\textwidth}
\centering
\includegraphics[width=0.45\textwidth]{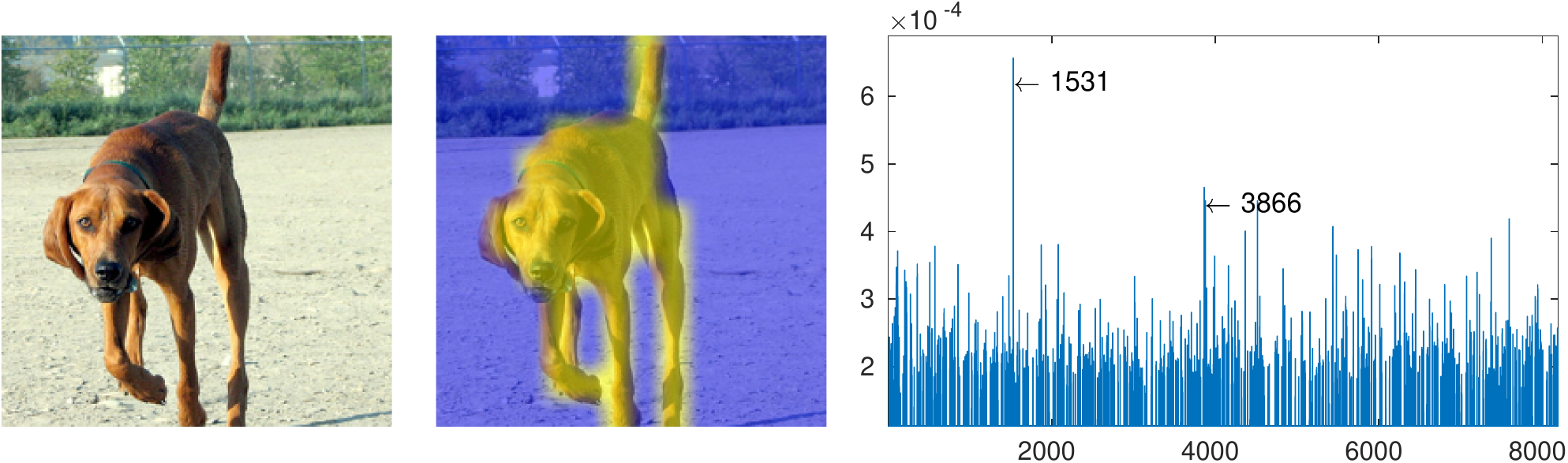}
\hspace{0.9cm}
\includegraphics[width=0.45\textwidth]{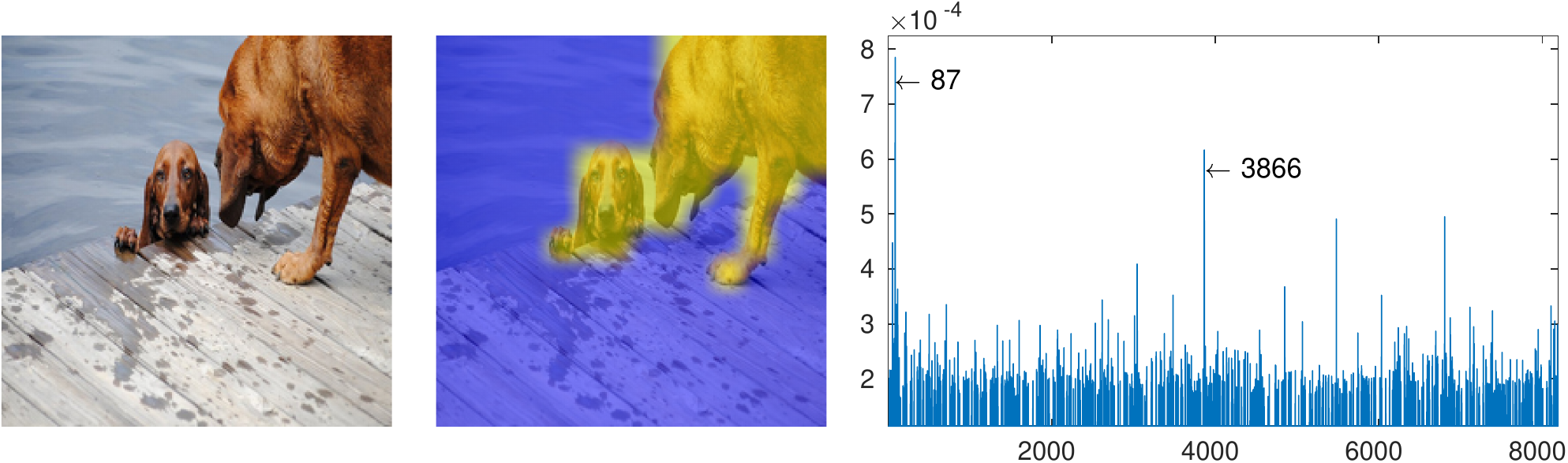}
\end{subfigure}
\begin{subfigure}[t]{\textwidth}
\centering
\includegraphics[width=0.45\textwidth]{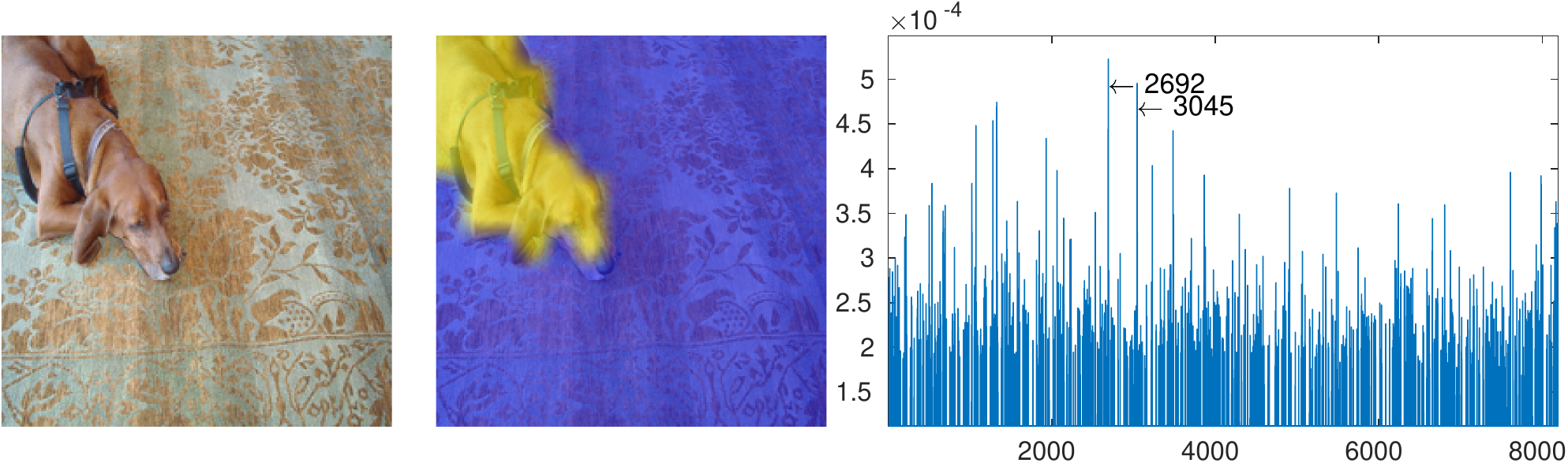}
\hspace{0.9cm}
\includegraphics[width=0.45\textwidth]{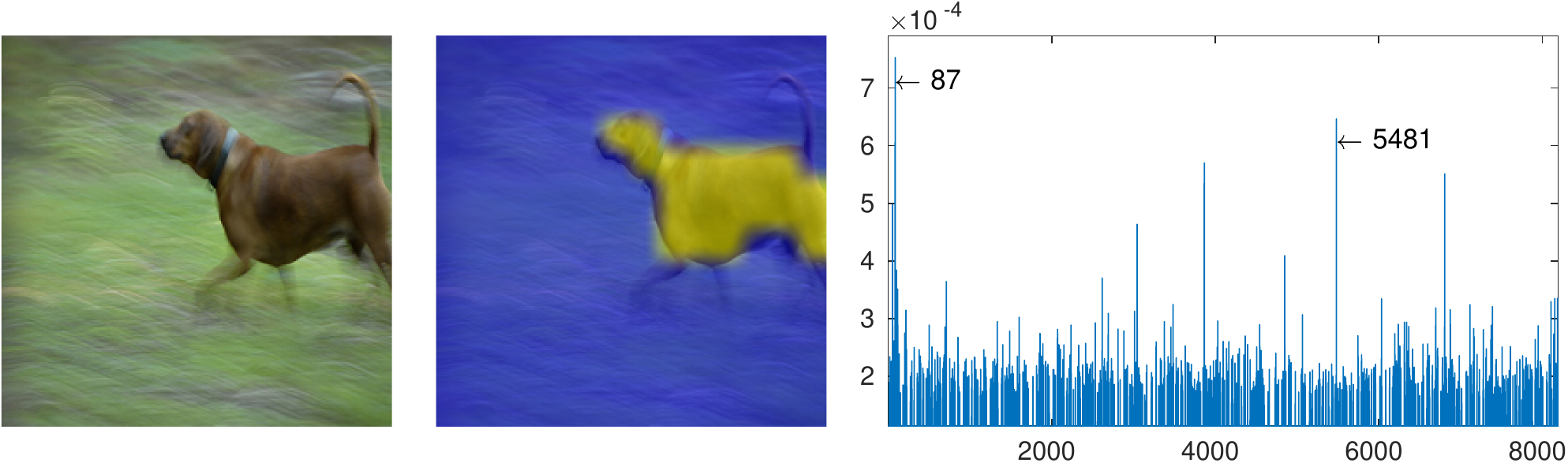}
\end{subfigure}
\begin{subfigure}[t]{\textwidth}
\centering
\includegraphics[width=0.45\textwidth]{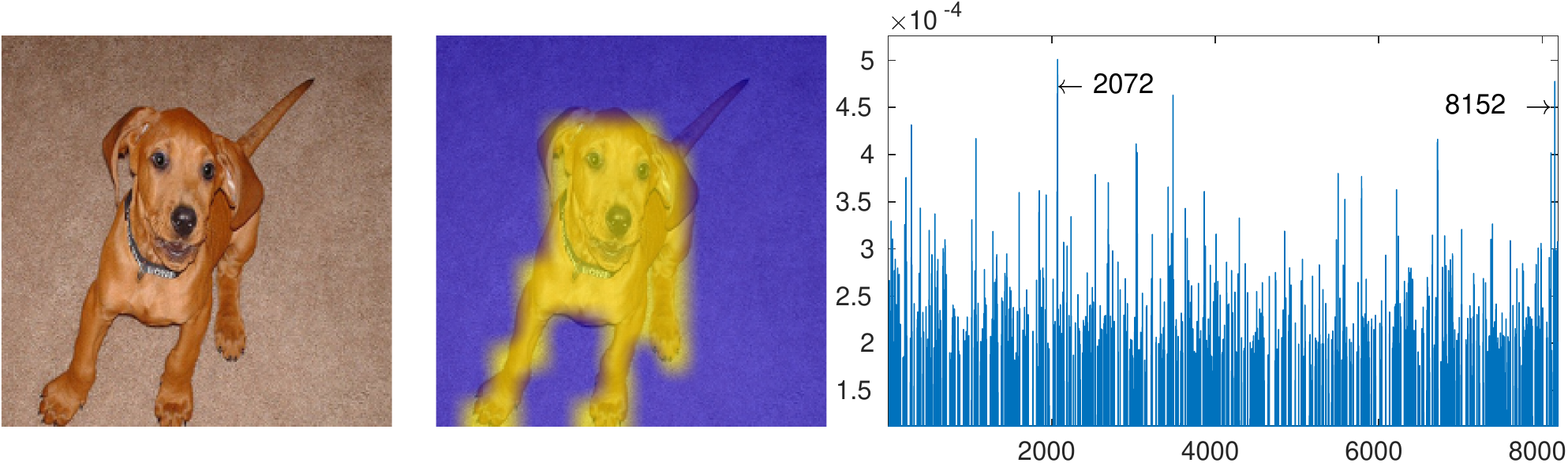}
\hspace{0.9cm}
\includegraphics[width=0.45\textwidth]{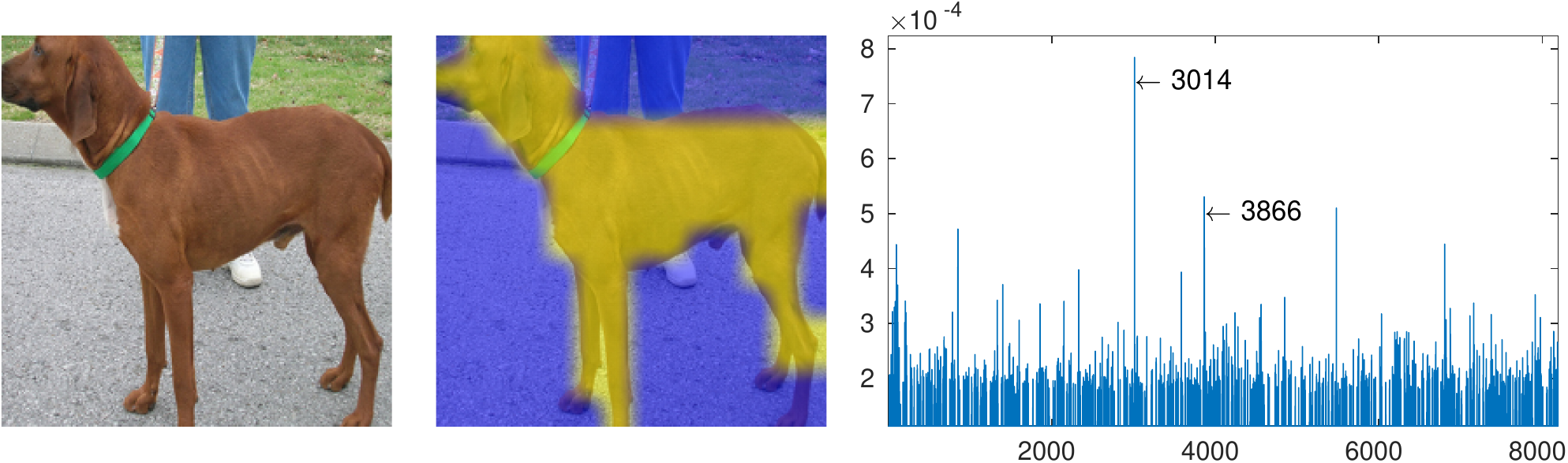}
\end{subfigure}
\begin{subfigure}[t]{\textwidth}
\centering
\includegraphics[width=0.45\textwidth]{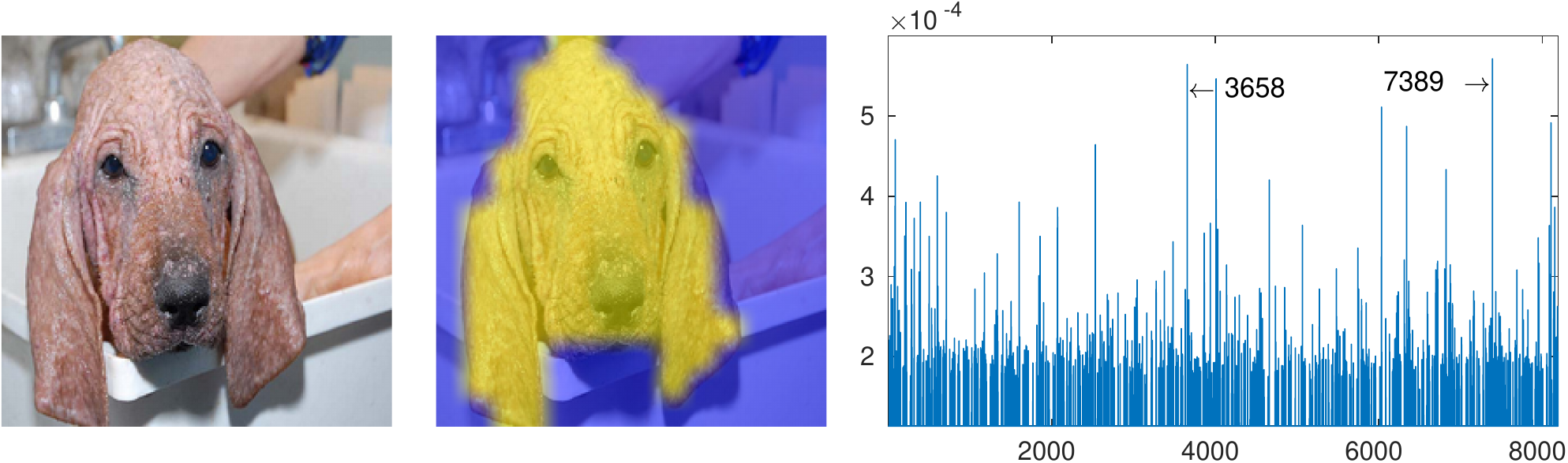}
\hspace{0.9cm}
\includegraphics[width=0.45\textwidth]{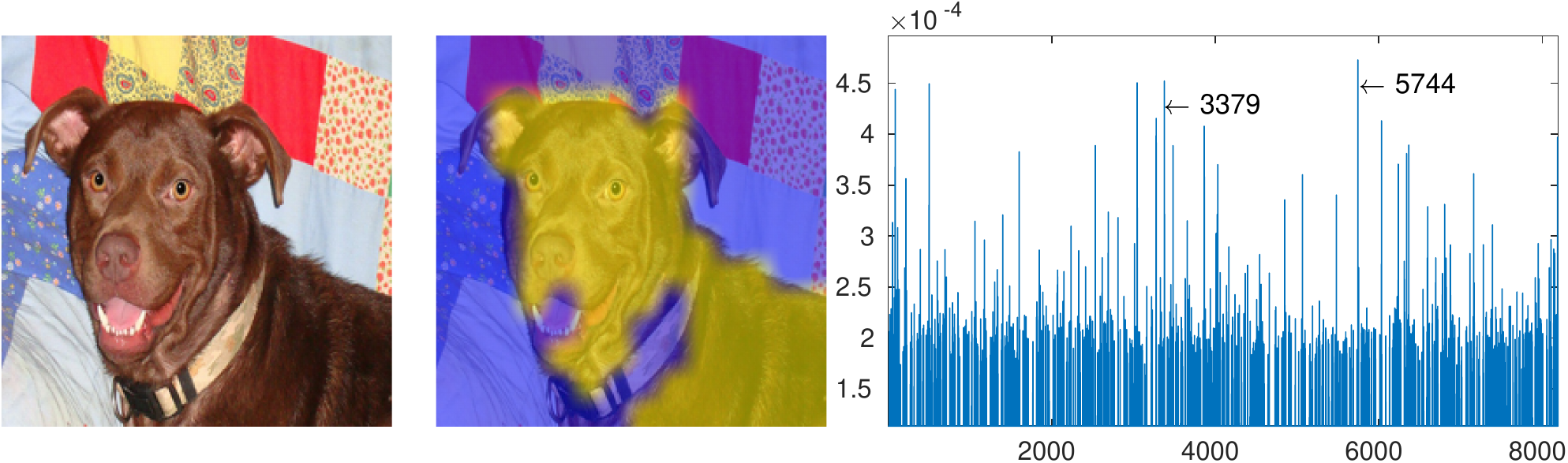}
\end{subfigure}
\begin{subfigure}[t]{\textwidth}
\centering
\includegraphics[width=0.3\textwidth]{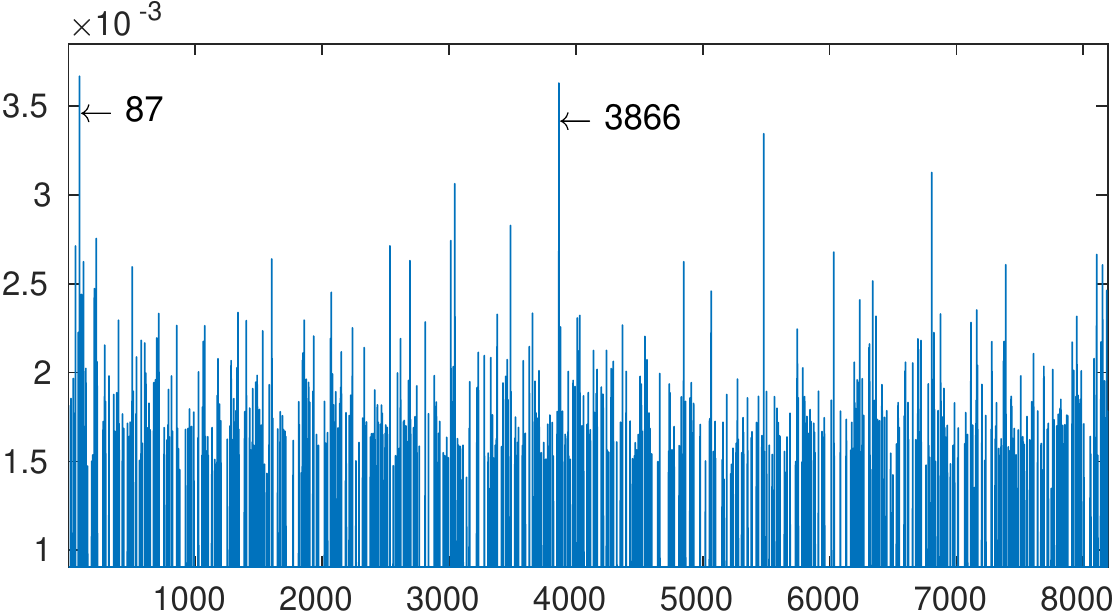}
\end{subfigure}
\caption{The image is clustered into two concepts which are self learnt. First self-learnt cluster is attending the dominant object (shown by yellow colour in centre column) and other is focusing on the environment (shown by blue colour in centre column). 
The bar plots are showing the probability of the learnt patch concepts corresponding to the dominant object. As can be seen the learnt patch concepts are consist across different instances of semantic concept in different images enforcing the thesis of MC-SSL0.0.}
\end{figure*}

\end{document}